\documentclass[10pt,twocolumn,letterpaper]{article}

\usepackage{iccv}
\usepackage{times}

\usepackage{graphicx}
\usepackage{amsmath}
\usepackage{amssymb}
\usepackage{booktabs}
\usepackage{multirow}
\usepackage{color}
\usepackage[table,dvipsnames]{xcolor}
\usepackage{pifont}
\usepackage{comment}
\usepackage{courier}
\usepackage[linesnumbered,ruled,vlined]{algorithm2e}
\usepackage{algpseudocode}
\SetKwInput{KwInput}{Input} 
\newcommand{\xmark}{\ding{55}}
\usepackage{bm}
\usepackage{tikz}
\usepackage{enumitem}
\usepackage{tabularx}
\usepackage{colortbl}
\usepackage{stmaryrd}
\usepackage{flushend}
\usepackage{makecell}
\usepackage{xcolor}
\usepackage{tabstackengine}
\usepackage{cite}
\usepackage{twemojis}
\usepackage[accsupp]{axessibility}

\usetikzlibrary{arrows.meta}

\usepackage[pagebackref,breaklinks,colorlinks,citecolor=RoyalBlue]{hyperref}
\usepackage[T1]{fontenc}
\usepackage{soul}

\usepackage[capitalize]{cleveref}

\crefname{section}{Sec.}{Secs.}
\Crefname{section}{Section}{Sections}
\Crefname{table}{Table}{Tables}
\crefname{table}{Tab.}{Tabs.}

\newcommand{\EncI}{E_\text{img}}
\newcommand{\EncT}{E_\text{txt}}

\newcommand{\mb}[1]{\ensuremath{\mathbf{#1}}}
\newcommand{\mc}[1]{\ensuremath{\mathcal{#1}}}
\newcommand{\bs}[1]{\ensuremath{\boldsymbol{#1}}}

\newcommand{\src}[1]{{#1}_\text{s}}
\newcommand{\trg}[1]{{#1}_\text{t}}
\newcommand{\stot}[1]{{#1}_{\text{s}\shortrightarrow\text{t}}}

\newcommand{\emb}[1]{\bar{#1}}

\newcommand{\var}[1]{{\small\,$\pm$#1}}
\newcommand{\vartn}[1]{{\scriptsize\,$\pm$#1}}
\newcommand{\query}[1]{\texttt{#1}}

\newcommand{\prompt}{\mathsf{TrgPrompt}}
\newcommand{\promptFeat}{\mathsf{TrgEmb}}
\newcommand{\freeze}{\texttwemoji{2744}}

\definecolor{srccolor}{rgb}{0.0, 0.4375, 0.0}
\definecolor{trgcolor}{rgb}{0.5977, 0.0, 0.5977}
\definecolor{nightcolor}{rgb}{0.0, 0.0, 0.8}
\definecolor{snowcolor}{rgb}{0.2, 0.6, 1.0}
\definecolor{gamecolor}{rgb}{1.0, 0.48, 0.48}
\definecolor{grayprompt}{rgb}{0.82, 0.82, 0.82}
\definecolor{colorcell}{rgb}{0.819, 0.94, 0.956}

\definecolor{colrelprompt}{rgb}{0.0, 0.4, 0.6}
\definecolor{colirrprompt}{rgb}{0.8, 0.0, 0.0}

\newcommand{\relprompt}{{\color{colrelprompt!100}relevant prompt}}
\newcommand{\irrprompt}{{\color{colirrprompt!100}irrelevant prompt}}

\definecolor{codeblue}{rgb}{0.25,0.5,0.5}

\SetCommentSty{mycommfont}

\newcommand{\method}{P{\O}DA\xspace}
\newcommand{\DAsetting}[2]{{#1}$\shortrightarrow${#2}}

\newcommand{\dashrule}[1][black]{%
  \color{#1}\rule[\dimexpr.5ex-.2pt]{4pt}{.4pt}\xleaders\hbox{\rule{4pt}{0pt}\rule[\dimexpr.5ex-.2pt]{4pt}{.4pt}}\hfill\kern0pt%
}
\newcommand{\rulecolor}[1]{%
  \def\CT@arc@{\color{#1}}%
}

\def\iccvPaperID{6091}

\iccvfinalcopy
\begin{document}

\title{
    \method{}: Prompt-driven Zero-shot Domain Adaptation 
}

\author{Mohammad Fahes${^1}$ \quad Tuan-Hung Vu$^{1,2}$ \quad  Andrei Bursuc$^{1,2}$ \quad Patrick Pérez$^{1,2}$ \quad  Raoul de Charette$^{1}$
{}\\
$^1$ Inria \quad \quad \quad  $^2$ Valeo.ai\\
\url{https://astra-vision.github.io/PODA}
}

\maketitle
\thispagestyle{empty}

%%%%%%%%% ABSTRACT
\begin{abstract}
Domain adaptation has been vastly investigated in computer vision but still requires access to target images at train time, which might be intractable in some uncommon conditions. In this paper, we propose the task of `Prompt-driven Zero-shot Domain Adaptation', where we adapt a model trained on a source domain using only a general description in natural language of the target domain, i.e., a prompt. First, we leverage a pretrained contrastive vision-language model (CLIP) to optimize affine transformations of source features, steering them towards the target text embedding while preserving their content and semantics. To achieve this, we propose Prompt-driven Instance Normalization (PIN). Second, we show that these prompt-driven augmentations can be used to perform zero-shot domain adaptation for semantic segmentation. Experiments demonstrate that our method significantly outperforms CLIP-based style transfer baselines on several datasets for the downstream task at hand, even surpassing one-shot unsupervised domain adaptation.
A similar boost is observed on object detection and image classification.
 The code is available at \,\emph{\url{https://github.com/astra-vision/PODA}~.
}

\end{abstract}

%%%%%%%%% BODY TEXT

\section{Introduction}
\label{sec:intro}

The last few years have witnessed tremendous success of supervised semantic segmentation methods towards better high-resolution predictions~\cite{long2015fully,chen2017deeplab,chen2018encoder,cheng2020panoptic, wang2020deep},
multi-scale processing~\cite{zhao2017pyramid, lin2017feature} or computational efficiency~\cite{zhao2018icnet}. In controlled settings where segmentation models are trained using data from the targeted operational design domains, the accuracy can meet the high industry-level expectations on in-domain data; yet, when tested on out-of-distribution data, these models often undergo drastic performance drops~\cite{ovadia2019can}.
This hinders their applicability in real-world scenarios for critical applications like in-the-wild autonomous driving.

\begin{figure}[t]
    \centering
    \includegraphics[width=1.0\linewidth]{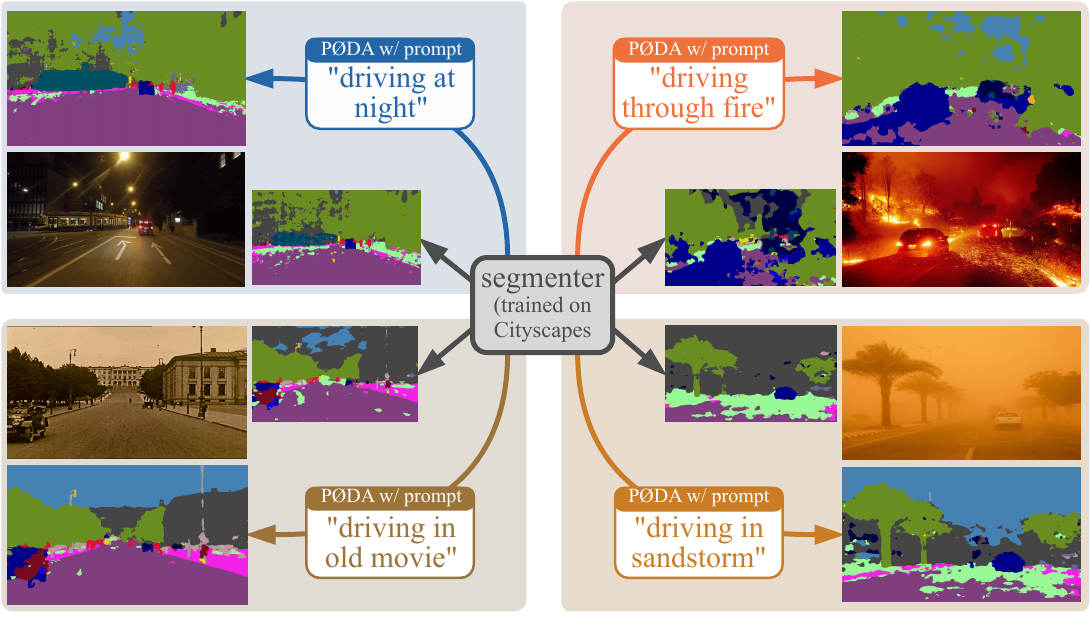}
     \caption{\textbf{Zero-shot adaptation with prompt.} \method{} enables the adaptation of a segmenter model (here, DeepLabv3+ \textit{trained on the source dataset Cityscapes}) to unseen conditions with only a prompt. Source-only predictions are shown as smaller segmentation masks to the left or right of the test images. 
     }
\label{fig:teaser}
\end{figure}

\begin{figure*}[t!]
    \centering
    \includegraphics[width=1.0\linewidth]{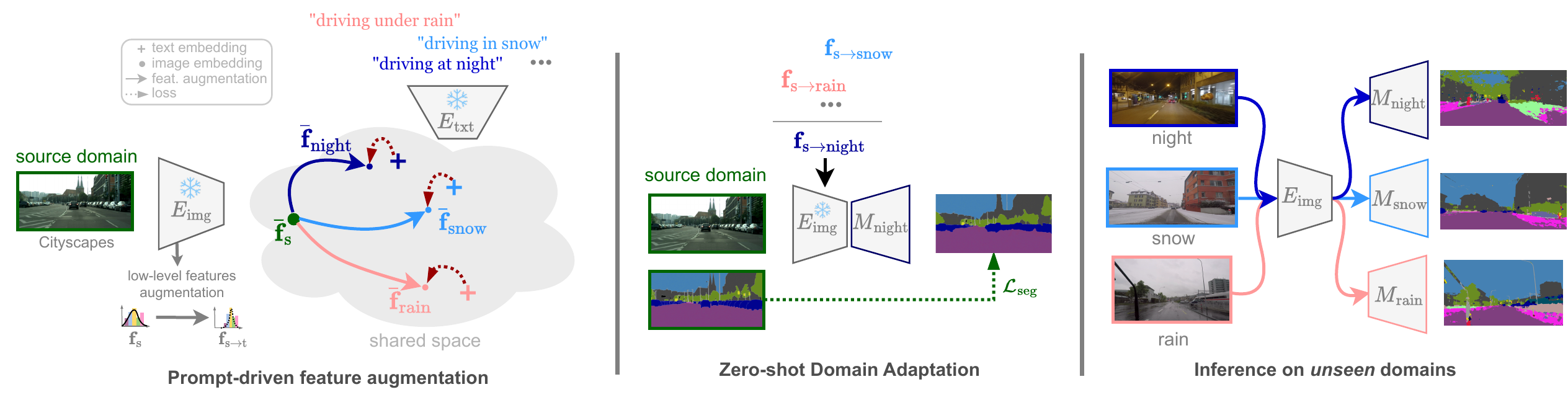}
    \vspace{-0.2cm}
    \caption{\textbf{Overview of \method{}, a Prompt-driven Zero-shot Domain Adaptation.} 
    (\textit{Left})~Using only a single textual description (``...'') of an \textit{unseen} target domain, we leverage a frozen ResNet encoder with CLIP weights to learn $\text{source}{\shortrightarrow}\text{target}$ low-level feature stylizations. 
 Applied to a source image low-level feature map $\src{\mb{f}}$ with latent embedding $\color{srccolor}\bar{\mb{f}}_{\text{s}}$, these stylizations provide augmented features $\mb{f}_{\text{s}\shortrightarrow\text{t}}$ which embeddings (here, $\color{nightcolor}\bar{\mb{f}}_{\text{night}}$,  $\color{snowcolor}\bar{\mb{f}}_{\text{snow}}$,  $\color{gamecolor}\bar{\mb{f}}_{\text{rain}}$) are closer to their respective target prompt embeddings ({\large \textbf{+}}). 
	(\textit{Middle})~Zero-shot domain adaptation is achieved by fine-tuning a segmenter model ($M$) on the feature-augmented source domain with the learned transformations. 
	(\textit{Right})~This enables inference on unseen domains.
	}
    \vspace{-0.2cm}
\label{fig:overview}
\end{figure*}

To mitigate this domain-shift problem~\cite{ben2010theory}, unsupervised domain adaptation (UDA)~\cite{ganin2016domain,sun2016deep,hoffman2018cycada,zou2018unsupervised,tsai2018learning,vu2019advent} has emerged as a promising solution.
Training of UDA methods requires labeled data from \emph{source} domain and unlabeled data from \emph{target} domain.
Though seemingly effortless, for some conditions even collecting unlabeled data is complex.
{For example, as driving through fire or sandstorm rarely occurs in real life, collecting raw data in such conditions is non-trivial.
One may argue on using Internet images for UDA.
However, in the industrial context, the practice of using public data is limited or forbidden.}
Recent works aim to reduce the burden of target data collection campaigns by devising one-shot~\cite{luo2020adversarial,wu2022style} UDA methods, \ie, using one target image for training.
Pushing further this line of research, we frame the challenging new task of prompt-driven zero-shot domain adaptation where
given a target domain description in natural language (\ie, a \textit{prompt}), our method accordingly adapts the segmentation model to this domain of interest. 
{Figure~\ref{fig:teaser} outlines the primary goal of our work with a few qualitative examples.
Without seeing any fire or sandstorm images during training, the adapted models succeed in segmenting out critical scene objects, exhibiting fewer errors than the original source-only model.\\}
\indent Our method, illustrated in \cref{fig:overview}, is made possible by leveraging the vision-language connections from the seminal CLIP model \cite{radford2021learning}. 
Trained on 400M web-crawled image-text pairs, CLIP has revolutionized multi-modal representation learning, bringing outstanding capability to zero-shot image synthesis~\cite{kwon2022clipstyler,gal2021stylegan,patashnik2021styleclip}, zero-shot multi-modal fusion~\cite{jatavallabhula2023conceptfusion}, zero-shot semantic segmentation~\cite{li2022language, zhou2022extract}, open-vocabulary object detection~\cite{minderer2022simple}, few-shot learning~\cite{cohen2022my}, etc. 
In our work, we exploit the CLIP's latent space and propose a simple and effective feature stylization mechanism that converts source-domain \textit{features} into target-domain ones (\cref{fig:overview}, left), which can be seen as a specific form of data \textit{augmentation}. 
Fine-tuning the segmentation model on these zero-shot synthesized features (\cref{fig:overview}, middle) helps mitigating the distribution gap between the two domains 
thus improving performance on unseen domains (\cref{fig:overview}, right). 
Owing to the standard terminology of ``prompt'' that designates the input text in CLIP-based image generation, we coin our approach \textit{Prompt-driven Zero-shot Domain Adaptation}, \method{} in short.

\noindent{}To summarize, our contributions are as follows:
\begin{itemize}[leftmargin=15pt,topsep=0pt,itemsep=0pt,parsep=0pt,partopsep=0pt]
    \item We introduce the novel task of prompt-driven zero-shot domain adaptation, which aims at adapting a source-trained model on a target domain provided \emph{only} {an arbitrary} textual description of the latter.
    \item Unlike other CLIP-based methods that navigate CLIP latent space using direct image representations, we alter only the features, without relying on the appearance in pixel space. We argue that this is particularly useful for downstream tasks such as semantic segmentation where good features are decisive (and sufficient). We present a simple and effective 
    \textit{Prompt-driven Instance Normalization (PIN)} layer to augment source features, where affine transformations of low-level features are optimized such that the representation in CLIP latent space matches the one of target-domain prompt.
    \item We show the versatility of our method by adapting source-trained semantic segmentation models to different conditions: (i) from clear weather/daytime to adverse conditions (snow, rain, night), (ii) from synthetic to real, (iii) from real to synthetic.
    Interestingly, \method{}  
    outperforms state-of-the-art one-shot unsupervised domain adaptation {without using any target image}.
    \item We show that \method{} 
    can also be applied to object detection and image classfication.
\end{itemize}

\section{Related works}

\smallskip\noindent\textbf{Unsupervised Domain Adaptation.~}
The UDA literature is vast and encompasses different yet connected approaches: adversarial learning~\cite{ganin2016domain,tsai2018learning}, self-training~\cite{zou2019confidence,li2019bidirectional}, entropy minimization~\cite{vu2019advent,pan2020unsupervised}, generative-based adaptation~\cite{hoffman2018cycada}, etc. The domain gap is commonly reduced at the level of the input~\cite{hoffman2018cycada,yang2020fda}, of the features~\cite{ganin2016domain,sun2016deep,wang2017deep,long2018conditional} or of the output~\cite{tsai2018learning,vu2019advent,pan2020unsupervised}.

Recently, the more challenging setting of One-Shot Unsupervised Domain Adaptation (OSUDA) has been proposed. To the best of our knowledge, two works on OSUDA for semantic segmentation exist~\cite{luo2020adversarial,wu2022style}. 
Luo et al.~\cite{luo2020adversarial} show that traditional UDA methods fail when only a single unlabeled target image is available.
To mitigate the risk of over-fitting on the style of the single available image, 
the authors propose a style mining algorithm, based on both a stylized image generator and a task-specific module.
Wu et al.~\cite{wu2022style} introduce an approach based on style mixing and patch-wise prototypical matching (SM-PPM).
During training, channel-wise mean and standard deviation of a randomly sampled source image's features  are linearly mixed with the target ones.
Patch-wise prototypical matching helps overcome negative adaptation~\cite{li2020content}.

In the more challenging zero-shot setting (where no target image is available), Lengyel et al.~\cite{lengyel2021zero} tackle day-to-night domain adaptation using physics priors. They introduce a color invariant convolution layer (CIConv) that is added to make the network invariant to different lighting conditions. We note that this zero-shot adaption 
is orthogonal to ours and restricted to a specific type of domain gap.

\smallskip\noindent\textbf{Text-driven image synthesis.~}
Recently, contrastive image-language pretraining has shown unprecedented success for multimodal learning in several downstream tasks such as zero-shot classification~\cite{radford2021learning}, multi-modal retrieval~\cite{jia2021scaling} and visual question answering~\cite{li2021align}.
This encouraged the community to modify images using text descriptions, a task that was previously challenging due to the gap between vision and language representations.
For example, StyleCLIP~\cite{patashnik2021styleclip} uses prompts to optimize StyleGAN~\cite{karras2019style} latent vectors and guide the generation process.
However, the generation is limited to the training distribution of StyleGAN.
To overcome this issue, StyleGAN-NADA~\cite{gal2021stylegan} utilizes CLIP embeddings of text-prompts to perform domain adaptation of the generator, which is in this case trainable. Similarly, for text-guided semantic image editing, FlexIT~\cite{couairon2022flexit} optimizes the latent code in VQGAN autoencoder's\cite{esser2021taming} space. 

For text-guided style transfer, CLIPstyler~\cite{kwon2022clipstyler} does not rely on a generative process. This setting is more realistic for not being restricted to a specific distribution, and challenging at the same time for the use of the encapsulated information in CLIP latent space. Indeed, there is no one-to-one mapping between image and text representations and regularization is needed to extract the useful information from a text embedding. Thus, in the same work~\cite{kwon2022clipstyler}, a U-net autoencoder that preserves the content is optimized while the output image embedding in CLIP latent space is varying during the optimization process.

We note that a common point in prior works is the mapping from pixel-space to CLIP latent space during the optimization process.  
In contrast with this, we directly manipulate deep features of the pre-trained CLIP visual encoder. 

\section{Prompt-driven Zero-shot Adaptation}

Our framework, illustrated in \cref{fig:overview}, builds upon CLIP~\cite{radford2021learning}, a vision-language model pre-trained on 400M image-text pairs crawled from the Internet.
CLIP trains jointly an image encoder~$\EncI$ and a text encoder~$\EncT$ over many epochs and learns an expressive representation space that effectively bridges the two modalities.
In this work, we leverage this property
to ``steer'' features from any source image towards a target domain in the CLIP latent space, 
with guidance from {an arbitrary} prompt describing the target domain, \eg, ``driving at night'' {or ``navigating the roads in darkness'' for the night domain}.
Our goal is to modify the style of source image features, bringing them closer to imaginary counterparts in the targeted domain (\cref{fig:overview}, left), while preserving their semantic content. 
The learned augmentations can then be applied on source images to generate features, in a zero-shot fashion, that correspond to the unseen target domain and can be further used to fine-tune the model towards handling target
domain (\cref{fig:overview}, middle). This ultimately allows inference on unseen domains only described by a simple prompt 
at train time (\cref{fig:overview}, right).

Our approach faces several challenges:
\emph{(i)} How to generate informative features for the target domain without having access to any image from it; 
\emph{(ii)} How to preserve pixel-wise semantics while augmenting features;
\emph{(iii)} Based on such features, how to adapt the source model to the \emph{unseen} target domain.
We address these questions in the following.

\smallskip\noindent\textbf{Problem formulation.~}
Our main task is semantic segmentation, that is, pixel-wise classification of input image into semantic segments.
We start from a $K$-class segmentation model $M$, pre-trained on a source domain dataset $\src{\mc{D}}=\{(\src{\mb{x}}, \src{\mb{y}}) \mid \src{\mb{x}} \in\mathbb{R}^{H\times W \times 3}, \src{\mb{y}} \in \{0,1\}^{H\times W \times K}\}$. By using a single predefined prompt $\prompt$ describing the targeted domain, we adapt the model $M$ such that its performance on the unseen test target dataset \makebox{$\trg{\mc{D}}=\{\trg{\mb{x}} \mid \trg{\mb{x}} \in\mathbb{R}^{H\times W \times 3}\}$} is improved. 
The segmenter $M$ is a DeepLabv3+ model~\cite{chen2018encoder} with CLIP image encoder~$\EncI$ (\eg, ResNet-50) as the frozen feature extractor backbone $M_\text{feat}$ and a randomly initialized pixel classification head $M_\text{cls}$:  $M=(M_\text{feat}, M_\text{cls})$.
We train $M$ 
in a supervised manner for the semantic segmentation task on the source domain.
In order to preserve the compatibility of the encoder features with the CLIP latent space we keep $M_\text{feat}$ frozen and train only the pixel classifier $M_\text{cls}$. 
Interestingly, we empirically show in \cref{tab:baselines} that keeping the feature extractor $M_\text{feat}$ frozen also prevents overfitting to the source in favor of generalization.
From the extractor we remove the attention pooling head of $\EncI$ to keep the spatial information for the pixel classifier.
We denote $\mb{f}$ the intermediate features extracted by $M_\text{feat}$ and $\emb{\mb{f}}$ their corresponding CLIP embedding 
computed with the attention pooling layer of $\EncI$.
In \cref{fig:method} we illustrate the difference between $\mb{f}$ and $\emb{\mb{f}}$.

\begin{table}[b]

	\setlength{\tabcolsep}{0.03\linewidth}
	\centering
	\begin{tabular}{cccccc}
		\toprule
		$M_\text{feat}$\,\freeze & CS &  Night &  Snow &  Rain & GTA5 \\
		\midrule
		Yes & 66.82 & \textbf{18.31} & \textbf{39.28} & \textbf{38.20} & \textbf{39.59} \\
		No & \textbf{69.17} & 14.40 & 22.27 & 26.33 & 32.91\\
		\bottomrule
	\end{tabular}
        \smallskip\caption{
	\textbf{Segmentation with source-only trained models}. Performance (mIoU\,\%) on ``night'', ``snow'' and ``rain'' parts of ACDC~\cite{sakaridis2021acdc} \textit{validation set} and on a subset of $1000$ GTA5 images for models trained on Cityscapes (CS). `$M_\text{feat}$\texttwemoji{2744}':  frozen backbone.
	}
\label{tab:baselines}
\end{table}

\smallskip\noindent\textbf{Overview of the proposed method.~} Our solution is to 
mine styles using source-domain low level features set $\src{\mc{F}}{=}\{\src{\mb{f}} {\mid} \src{\mb{f}} {=} \texttt{feat-ext}(M_\text{feat}, \src{\mb{x}})\}$ and $\promptFeat$,
where $\promptFeat{=}\EncT(\prompt)$ is the CLIP text embedding of the target domain prompt.
For generality, $\texttt{feat-ext}(\cdot)$ can pull 
features from any desired layer but we later show that using the lowest features works best.

\begin{algorithm}[t]
	\small
	\SetAlgoLined
	\SetKwFunction{Mean}{mean}
	\SetKwFunction{Std}{std}
    \SetKwFunction{PIN}{PIN}
	\SetKwFunction{Ge}{get-embedding}
	\SetKwInOut{Input}{Input} 
	\SetKwInOut{Output}{Output} 
	\SetKwInOut{Parameter}{Param}
	
	\Input{Set $\src{\mc{F}}$ of source image features\\
		Target domain description embedding $\promptFeat$\\
	}
	\Parameter{Number $N$ of optimization steps\\
		Learning rate $lr$ and momentum $m$\\
            ~of gradient descend (\texttt{GD})}
	\Output{Set $\stot{\mc{S}}$ of 
 target styles} 
	
	$\stot{\mc{S}} \gets \emptyset$ \\
	\ForEach{$\src{\mb{f}} \in \src{\mc{F}}$}{%
		$\bs{\mu}^{0} \gets \Mean(\src{\mb{f}})$ \\
		$\bs{\sigma}^{0} \gets \Std(\src{\mb{f}})$ \\
		\tcp{Optimization}
		\For{$\text{i} = 1,2, \cdots, N$} {%
			$\mb{f}_{\text{s}\shortrightarrow\text{t}}^{i} \gets \PIN(\src{\mb{f}}, \bs{\mu}^{i-1}, \bs{\sigma}^{i-1})$ \\
			$\emb{\mb{f}}_{\text{s}\shortrightarrow\text{t}}^{i} \gets \Ge(\mb{f}_{\text{s}\shortrightarrow\text{t}}^{i})$\\ 
			$\bs{\mu}^{i} \gets \texttt{GD}^{lr}_{m}(\bs{\mu}^{i-1}, \nabla_{\bs{\mu}}\mc{L}_{\bs{\mu}, \bs{\sigma}}(\emb{\mb{f}}_{\text{s}\shortrightarrow\text{t}}^{i}, \promptFeat))$ \\ 
			$\bs{\sigma}^{i} \gets \texttt{GD}^{lr}_{m}(\bs{\sigma}^{i-1}, \nabla_{\bs{\sigma}}\mc{L}_{\bs{\mu}, \bs{\sigma}}(\emb{\mb{f}}_{\text{s}\shortrightarrow\text{t}}^{i}, \promptFeat))$
		}
    	   $(\bs{\mu}_{t}, \bs{\sigma}_{t}) \gets (\bs{\mu}^{N}, \bs{\sigma}^{N})$ \\
		$\stot{\mc{S}} \gets \stot{\mc{S}} \cup \{(\bs{\mu}_t, \bs{\sigma}_t)\}$
	}
	\caption{
		Style Mining (see Fig.~\protect{\ref{fig:method}})}
	\label{algo:style_mining}
\end{algorithm}

The $\texttt{augment}(\cdot)$ operation, depicted in \cref{fig:method}, augments the style-specific components of $\src{\mb{f}}$ with guidance from the target domain prompt, synthesizing $\stot{\mb{f}}$ with 
style information from the target domain.
We emphasize that 
the features $\src{\mb{f}}$ and $\stot{\mb{f}}$ have the same size $h\times w \times c$ and identical semantic content, though they encapsulate different visual styles.
For adaptation, the source features $\src{\mb{f}}$ are augmented with the mined styles then used to
fine-tune the classifier $M_\text{cls}$, resulting in the final adapted model. The overall pseudo-code is provided in Supplementary Material.

\subsection{Zero-shot Feature Augmentation}
\label{sec:feat_aug}

We take inspiration from Adaptive Instance Normalization (AdaIN)~\cite{huang2017arbitrary}, an elegant formulation for transferring style-specific components across deep features.
In AdaIN, the styles are represented by the channel-wise mean $\bs{\mu} \in \mathbb{R}^{c}$ and standard deviation $\bs{\sigma} \in \mathbb{R}^{c}$ of features, with $c$ the number of channels. Stylizing a source feature $\src{\mb{f}}$ with an arbitrary target style $(\mu(\trg{\mb{f}}), \sigma(\trg{\mb{f}}))$ 
reads:
\begin{equation}
	\small
	\texttt{AdaIN}(\src{\mb{f}}, \trg{\mb{f}}) = \sigma(\trg{\mb{f}}) \left( \frac{\src{\mb{f}} - \mu(\src{\mb{f}})}{\sigma(\src{\mb{f}})} \right) + \mu(\trg{\mb{f}}),
	\label{eqn:adain}
\end{equation}
with $\mu(\cdot)$ and $\sigma(\cdot)$ as the two functions returning channel-wise mean and standard deviation of input feature; multiplications and additions are element-wise.

\begin{figure}
	\centering
	\includegraphics[trim=2 0 0 0, clip,width=1\linewidth]{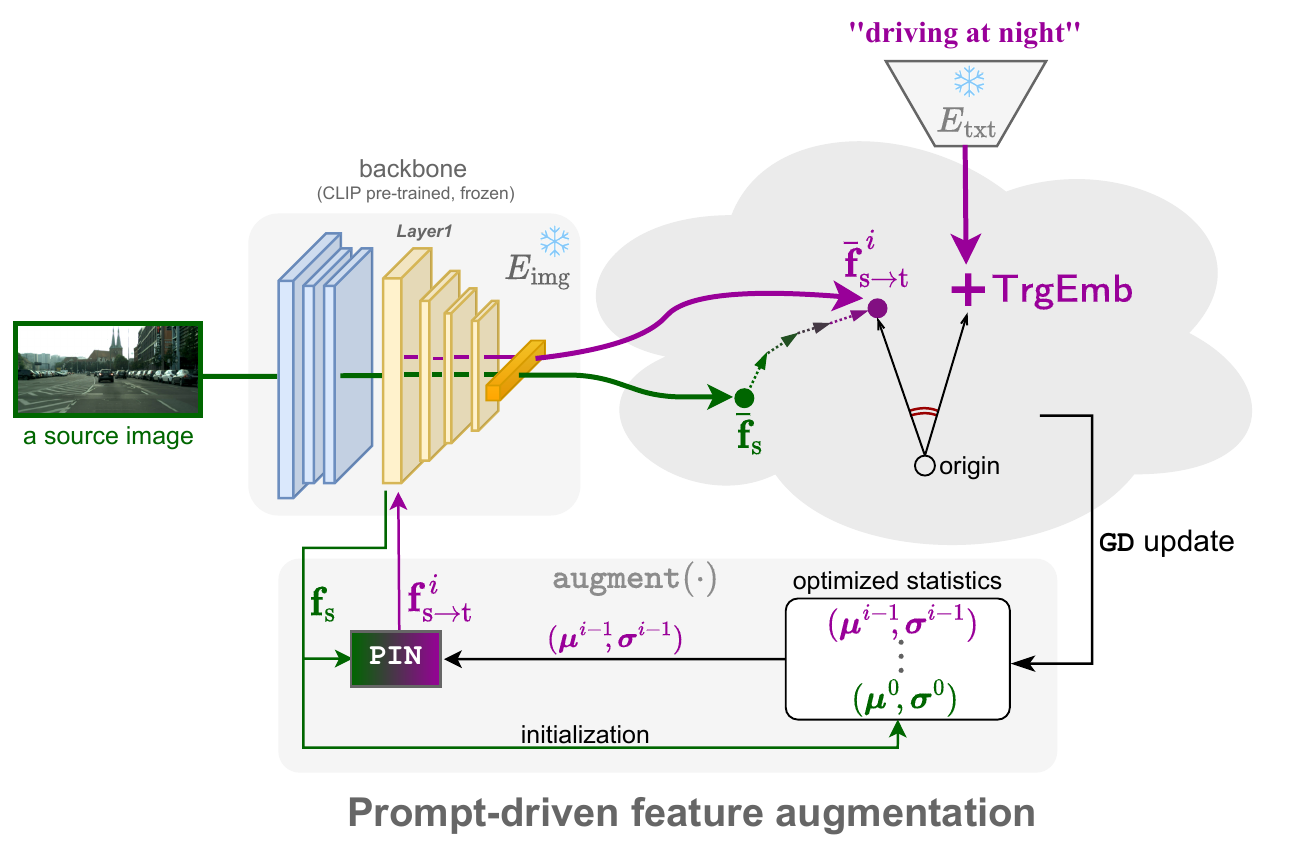}
	\vspace{-0.5cm}
	\caption{\textbf{Target style mining from a source image}. 
 We illustrate here the optimization loops of \cref{algo:style_mining}.
	The source image 
	is forwarded 
 through the CLIP image encoder $\EncI$ to extract low-level features $\color{srccolor} \src{\mb{f}}$ and subsequent CLIP embedding $\color{srccolor} \src{\emb{\mb{f}}}$.
	At each optimization step $i$, $\texttt{augment}(\cdot)$ takes the style of the previous iteration, 
	$\color{trgcolor}(\bs{\mu}^{i-1}\!\!,\bs{\sigma}^{i-1})$ and injects it within $\color{srccolor} \src{\mb{f}}$ via the \texttt{PIN} layer, 
    to synthesize~$\color{trgcolor}\textbf{f}^i_{\text{s}\shortrightarrow\text{t}}$ and the corresponding embedding~$\color{trgcolor}\bar{\textbf{f}}^i_{\text{s}\shortrightarrow\text{t}}$. 
    The loss $\color{BrickRed}\mc{L}_{\bs{\mu}, \bs{\sigma}}$ is the cosine distance between~$\color{trgcolor}\bar{\textbf{f}}^i_{\text{s}\shortrightarrow\text{t}}$ and the target prompt embedding~$\protect\promptFeat$. Its optimization via gradient descent updates style to $\textcolor{trgcolor}{(\bs{\mu}^{i}\!,\bs{\sigma}^{i})}$.}
	
\label{fig:method}
\end{figure}

We design our augmentation strategy 
around AdaIN as it can effectively manipulate the style information with a small set of parameters. In the following, we present our 
augmentation strategy which 
mines target styles. 

As we do not have access to any target (\ie style) image, $\mu(\trg{\mb{f}})$ and $\sigma(\trg{\mb{f}})$ are unknown. Thus, we propose \makebox{Prompt-driven} Instance Normalization (PIN):
\begin{equation}
	\small
	\texttt{PIN}(\src{\mb{f}}, \bs{\mu}, \bs{\sigma}) = \bs{\sigma} \left( \frac{\src{\mb{f}} - \mu(\src{\mb{f}})}{\sigma(\src{\mb{f}})} \right) + \bs{\mu},
	\label{eqn:PIN}
\end{equation}
where ${\bs{\mu}}$ and ${\bs{\sigma}}$ are optimizable variables driven by a prompt.

We aim to augment source image features $\src{\mc{F}}$ such that they capture the style of the target domain.
Here, the prompt describing a target domain could be fairly generic.
For instance, one can use prompts like
``driving at night'' or ``driving under rain'' 
to bring source features closer to the nighttime or rainy domains.
The prompt is processed by the CLIP text encoder $\EncT$ into the $\promptFeat$ embedding.

We describe in~\cref{algo:style_mining} the first step of our zero-shot feature augmentation procedure: mining the set $\stot{\mc{S}}$ of 
styles in targeted domain.
For each source feature map $\src{\mb{f}} \in \src{\mc{F}}$, we want to mine style statistics corresponding to an imaginary target feature map $\trg{\mb{f}}$. 
To this end, we formulate style mining as an optimization problem over the original source feature $\src{\mb{f}}$, \ie 
optimizing $(\bs{\mu}, \bs{\sigma})$ in~\cref{eqn:PIN}. 
The optimization objective is defined as the cosine distance 
in the CLIP latent space between the CLIP embedding $\emb{\mb{f}}_{\text{s}\shortrightarrow\text{t}}$ of the stylized feature 
$\stot{\mb{f}} = \texttt{PIN}(\src{\mb{f}}, \bs{\mu}, \bs{\sigma})$ and the description embedding $\promptFeat$ of target domain:
\begin{equation}
	\small
	\mc{L}_{\bs{\mu}, \bs{\sigma}}(\emb{\mb{f}}_{\text{s}\shortrightarrow\text{t}}, \promptFeat) = 1-\frac{\emb{\mb{f}}_{\text{s}\shortrightarrow\text{t}}\cdot\promptFeat}{\|\emb{\mb{f}}_{\text{s}\shortrightarrow\text{t}}\|\,\|\promptFeat\|}\,.
	\label{eqn:loss_dir}
\end{equation}
This CLIP-space cosine distance, already used in prior text-driven image editing works~\cite{patashnik2021styleclip}, 
aims to steer the stylized features in the direction of the target text embedding. 
One step of the optimization is illustrated in \cref{fig:method}. 
In practice, we run several such steps leading to the mined target style denoted $(\bs{\mu}_t, \bs{\sigma}_t)$.

As there might be a variety of styles in a target domain, our mining populates the $\stot{\mc{S}}$ set with as many variations of target style as there are source images, hence
$|\stot{\mc{S}}| = |\src{\mc{D}}|$.

Intuitively, our simple augmentation strategy can be seen as a cost-efficient way to cover the distribution of the target domain by starting from different anchor points in the CLIP latent space coming from the source images and steering them in the direction of the target text embedding.
This mitigates the diversity problem discussed in one-shot feature augmentation in~\cite{luo2020adversarial,wu2022style}.

\subsection{Fine-tuning for Adaptation}
\label{sec:fine_tune}
For adaptation, at each training iteration we stylize the source features using a mined target style $({\bs{\mu}}_t, \bs{\sigma}_t)$ randomly selected from $\stot{\mc{S}}$. The augmented features are computed as 
$\mb{f}_{\text{s}\shortrightarrow\text{t}} = 
 \texttt{PIN}(\src{\mb{f}}, \bs{\mu}_t, \bs{\sigma}_t)$ 
and are used for fine-tuning the classifier $M_\text{cls}$ of the segmenter $M$ (\cref{fig:overview}, middle). 
As we only adjust the feature style which keeps the semantic-content unchanged~\cite{huang2017arbitrary}, we can still use the labels ${\src{\mb{y}}}$ to train the classifier with a standard segmentation loss.
To this end, we simply forward augmented features through remaining layers in $M_\text{feat}$ followed by $M_\text{cls}$.
In the backward pass, only weights of $M_\text{cls}$ are updated by the loss gradients.
We denote the fine-tuned model as $M' = (M_\text{feat}, M'_\text{cls})$ and evaluate it on images with conditions and styles which were never seen during any of the training stages.

\section{\method{} for semantic segmentation}

\subsection{Implementation details}
\label{sec:exp_details}
We use the DeepLabv3+ architecture~\cite{chen2018encoder} with the backbone $M_\text{feat}$ initialized from the image encoder $\EncI$ of the pre-trained CLIP-ResNet-50 model\footnote{\url{https://github.com/openai/CLIP}}.

\smallskip\noindent\textbf{Source-only training.~}
{The network is trained for  
$200$k iterations on random $768 \times 768$ crops with batch size~$2$.}
We use a polynomial learning rate schedule with initial $lr{=}10^{-1}$ for the classifier {and $lr{=}10^{-4}$ for backbone when not frozen} (see~\cref{tab:baselines}). 
We optimize with Stochastic Gradient Descent~\cite{bottou2010large}, momentum $0.9$ and weight decay $10^{-4}$.
{We apply standard color jittering and horizontal flip to crops.}

\smallskip\noindent\textbf{Zero-shot feature augmentation.~}
For the feature augmentation step, we use the source feature maps after the first layer (\emph{Layer1}):  $\src{\mb{f}}\in \mathbb{R}^{192 \times 192 \times 256}$.
The style parameters $\bs{\mu}$ and $\bs{\sigma}$ are $256$D real vectors.
The CLIP embeddings are $1024$D vectors.
{We adopt the Imagenet templates from~\cite{radford2021learning} to encode the target descriptions in~$\prompt$.}

\smallskip\noindent\textbf{Classifier fine-tuning.~} Starting from the source-only pre-trained model, we fine-tune the classifier $M_\text{cls}$ on batches of $8$~augmented features $\stot{\mb{f}}$ 
for $2,000$ iterations.
Polynomial schedule is used with the initial $lr=10^{-2}$. We always use the last checkpoint for evaluation.

\smallskip\noindent\textbf{Datasets.~} 
As source, we use Cityscapes~\cite{cordts2016cityscapes}, composed of $2,975$ training and $500$ validation images featuring $19$ semantic classes. 
Though we adapt towards a prompt \textit{not} a dataset, we need adhoc datasets to test on.
We report main results using ACDC~\cite{sakaridis2021acdc} because it has urban images captured in adverse conditions.
We also study the applicability of \method~to the two settings of
\DAsetting{real}{synthetic} (Cityscapes as source, and evaluating on GTA5~\cite{richter2016playing}) and \DAsetting{synthetic}{real} (GTA5 as source, and evaluating on Cityscapes).
We evaluate on the validation set when provided, and for GTA5 evaluation we use a random subset of $1,000$ images.

\smallskip\noindent\textbf{Evaluation protocol.~} Mean Intersection over Union (mIoU\%) is used to measure adaptation performance. We test all models on target images at their original resolutions.
For baselines and \method{}, we always report the mean and standard deviation over five models trained with different random seeds.

\subsection{Main results}
\label{sec:main_res}
We consider the following adaptation scenarios: \DAsetting{day}{night},~\DAsetting{clear}{snow},~\DAsetting{clear}{rain},
~\DAsetting{real}{synthetic} and~\DAsetting{synthetic}{real}.
We report zero-shot adaption results of~\method~in the addressed set-ups, comparing against two state-of-the-art baselines: CLIPstyler~\cite{kwon2022clipstyler} for zero-shot style transfer and SM-PPM~\cite{wu2022style} for one-shot UDA.
Both~\method~and CLIPstyler models see no target images during training.
{In this study, we arbitrarily choose a simple prompt to describe each domain.
We show later in~\cref{sec:abla} more results using other relevant prompts with similar meanings -- showcasing that our adaptation gain is little sensitive to prompt selection.}
For SM-PPM, one random target image from the training set is used.

\smallskip\noindent\textbf{Comparison to CLIPstyler~\cite{kwon2022clipstyler}.~}
CLIPstyler is a style transfer method that {also} makes use of the pre-trained CLIP model {but} for zero-shot stylizing {of source images}.
We consider CLIPstyler\footnote{We use official code \url{https://github.com/cyclomon/CLIPstyler} and follow the recommended configs.} as the most comparable zero-shot baseline for \method~as both are built upon CLIP, though with different mechanisms and different objectives.
Designed for style transfer, CLIPstyler produces images that exhibit characteristic styles of the input text prompt.
However the stylized images 
{can have multiple} artifacts which hinder their usability in the downstream segmentation task.
This is visible in~\cref{fig:clipstyler} which shows stylized examples from CLIPstyler with \method target prompts. 
{Zooming in, we note that stylization of snow or game added snowy roads or Atari game \textit{on the buildings}, respectively.}

Starting from source-only model, we fine-tune the classifier on stylized images, as similarly done in \method{} {with the augmented features}. Table \ref{tab:main_results} compares \method~against the source-only model and CLIPstyler.~\method~consistently {outperforms the two baselines.}
CLIPstyler brings
some improvements over source-only in~\DAsetting{Cityscapes}{Night} and \DAsetting{Cityscapes}{Snow}. 
In other scenarios, \eg, rain, {CLIPstyler even performs worse than source-only.}

\begin{table}[t!]

		\setlength{\tabcolsep}{0.04\linewidth}
		\centering
		\resizebox{0.98\linewidth}{!}{
                \newcolumntype{H}{>{\setbox0=\hbox\bgroup}c<{\egroup}@{}}
			\begin{tabular}{lllc}
			\toprule
			Source
			& Target eval.
			& Method
			& mIoU[\%]\\
			\midrule
        \multirow{16}{*}{CS}& \multicolumn{3}{c}{\cellcolor{gray!34}$\prompt$ = ``driving at night''}\\
			& \multirow{3}{*}{ACDC Night} & source-only & 18.31\textcolor{white}{\var{0.00}}\\
    & & CLIPstyler & 21.38\var{0.36} \\
    & & \method & \textbf{25.03}\var{0.48} \\
   
        & \multicolumn{3}{c}{\cellcolor{gray!34}$\prompt$ = ``driving in snow''}\\
        & \multirow{3}{*}{ACDC Snow} & source-only &  39.28\textcolor{white}{\var{0.00}}\\
    & & CLIPstyler & 41.09\var{0.17} \\
        & & \method & \textbf{43.90}\var{0.53} \\

        &\multicolumn{3}{c}{\cellcolor{gray!34}$\prompt$ = ``driving under rain''}\\
		& \multirow{3}{*}{ACDC Rain} & source-only & 38.20\textcolor{white}{\var{0.00}}\\
  & & CLIPstyler & 37.17\var{0.10} \\
  & & \method &  \textbf{42.31}\var{0.55}\\

    &\multicolumn{3}{c}{\cellcolor{gray!34}$\prompt$ = ``driving in a game''}\\
    & \multirow{3}{*}{GTA5} & source-only & 39.59\textcolor{white}{\var{0.00}}\\
    & & CLIPstyler & 38.73\var{0.16} \\
    & & \method & \textbf{41.07}\var{0.48}\\
	\arrayrulecolor{black}		
	\midrule
	\multirow{4}{*}{GTA5} &\multicolumn{3}{c}{\cellcolor{gray!34}$\prompt$ = ``driving''}\\
    & \multirow{3}{*}{CS} & source-only &  36.38\textcolor{white}{\var{0.00}} \\
    & & CLIPstyler &  31.50\var{0.21} \\
    & & \method & \textbf{40.08}\var{0.52} \\
    \bottomrule
    \end{tabular}}
    \smallskip\caption{\textbf{Zero-shot domain adaptation in semantic segmentation.} Performance (mIoU\%) of~\method~compared against CLIPstyler~\cite{kwon2022clipstyler} and source-only baseline. Results are grouped by source domain and {\setlength{\fboxsep}{1pt}\colorbox{gray!34}{$\prompt$}}. CS stands for Cityscapes~\cite{cordts2016cityscapes}. The $\prompt$s are simply chosen, not engineered.}
   
	\label{tab:main_results}
\end{table}

\begin{figure}[t!]

	\newcommand{\rowqual}[1]{ %
	    \includegraphics[width=0.19\linewidth]{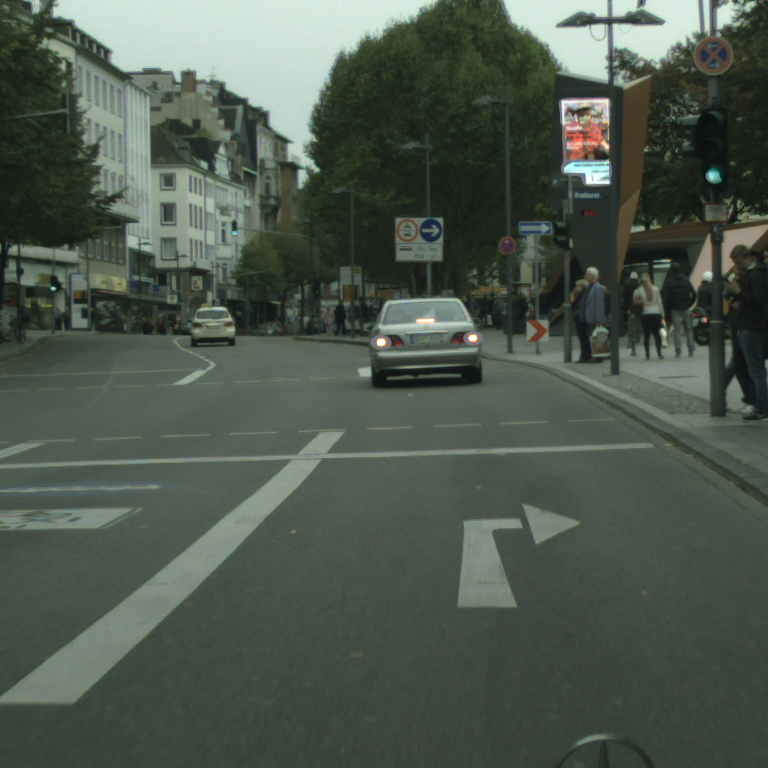}&
		\includegraphics[width=0.19\linewidth]{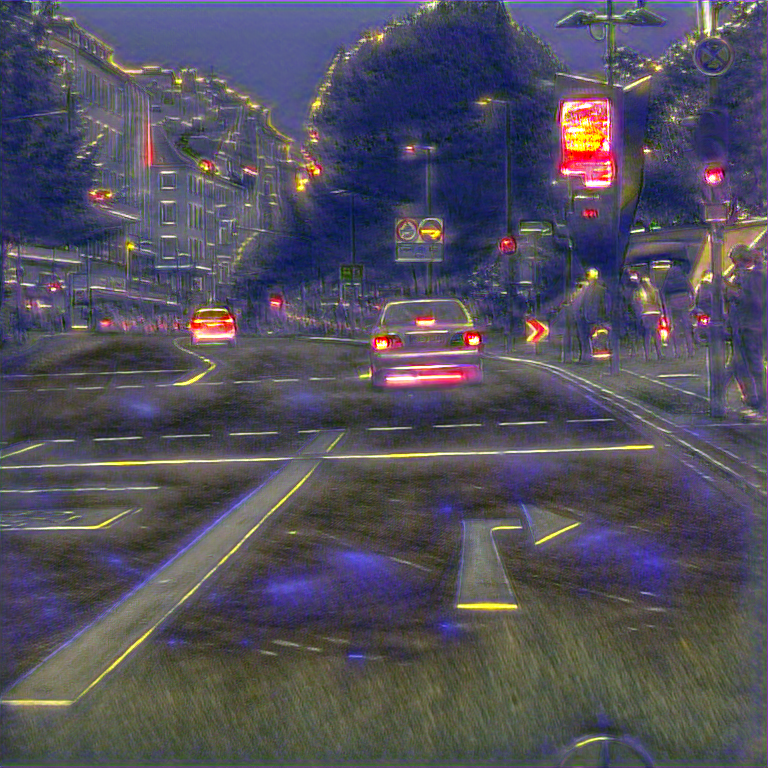}&
		\includegraphics[width=0.19\linewidth]{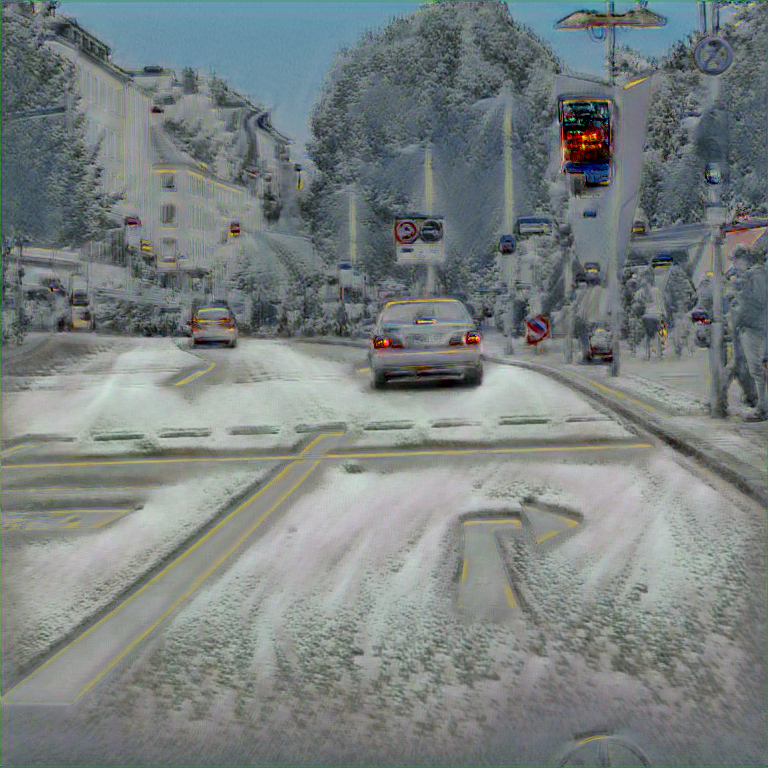}&
		\includegraphics[width=0.19\linewidth]{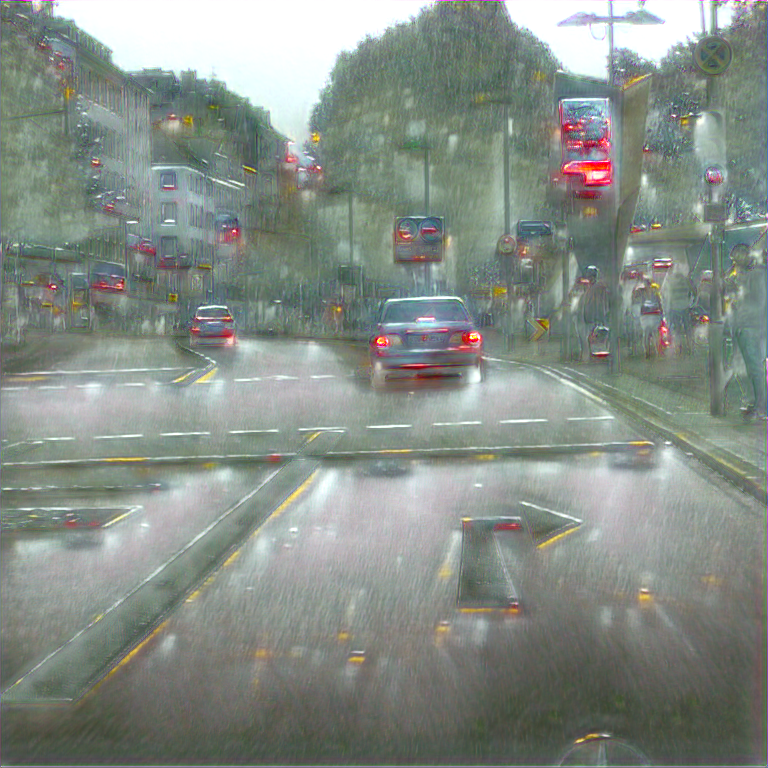}&
		\includegraphics[width=0.19\linewidth]{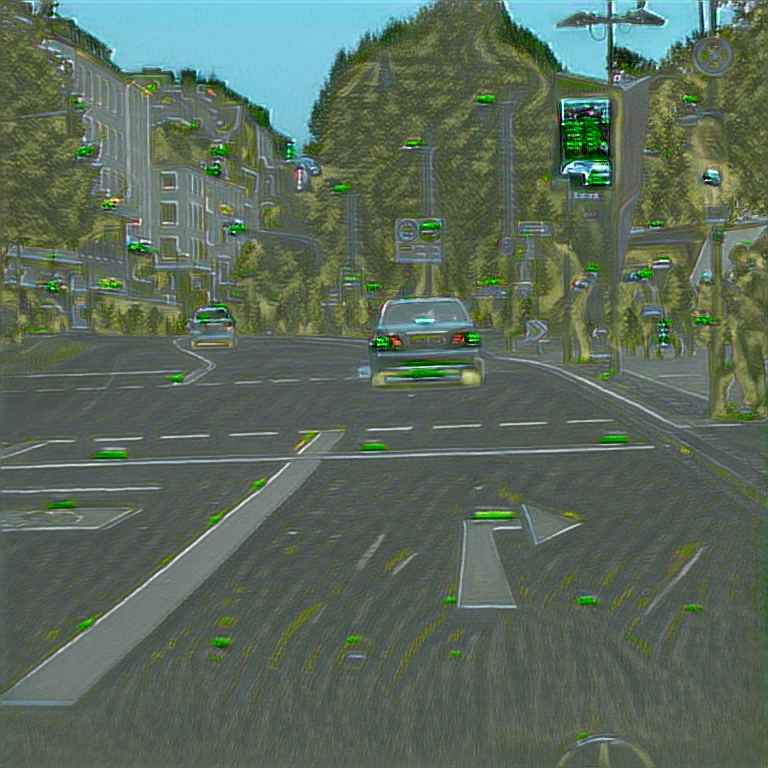}
	}
	\small	
	\setlength{\tabcolsep}{0.002\linewidth}
	\centering
	\begin{tabular}{cccccc}
				& Cityscapes & night & snow & rain & game\\
		\rotatebox{90}{} & \rowqual{CLIPstyler_visuals/33}\\ 
	\end{tabular}
    \smallskip
    \caption{\textbf{CLIPstyler~\cite{kwon2022clipstyler} stylization.} A sample Cityscapes image stylized using {adhoc target prompts}. {Translated images exhibit visible artifacts, potentially harming adaptation, \eg rain in~\cref{tab:main_results}}}
    \label{fig:clipstyler}
\end{figure}

\begin{figure}[t]
	\newcommand{\rowqual}[1]{ %
		\includegraphics[width=0.192\linewidth]{figures/qualitative/#1image.png}&
		\includegraphics[width=0.192\linewidth]{figures/qualitative/#1target.png}&
		\includegraphics[width=0.192\linewidth]{figures/qualitative/#1pred_src.png}&
		\includegraphics[width=0.192\linewidth]{figures/qualitative/#1pred_ours.png}
	}
	\scriptsize	
	\setlength{\tabcolsep}{0.002\linewidth}
	\centering
        \newcolumntype{H}{>{\setbox0=\hbox\bgroup}c<{\egroup}@{}}
    \resizebox{1.0\linewidth}{!}{%
        \tiny
	\begin{tabular}{ccccc}
				& {\scriptsize input} & {\scriptsize ground truth} & {\scriptsize source-only} & {\scriptsize\method}\\
		\rotatebox{90}{\,\,\DAsetting{CS}{Night}} & \rowqual{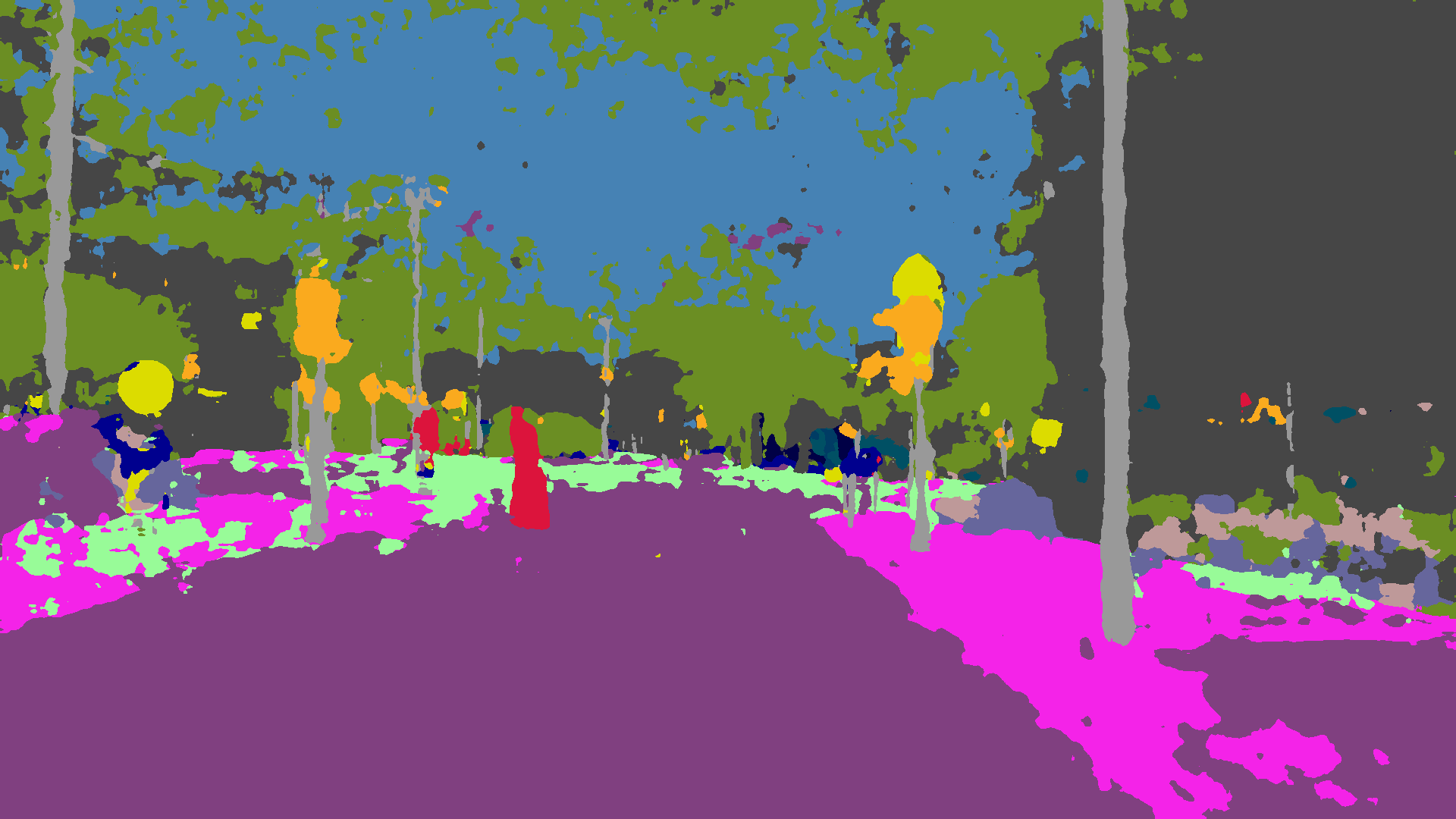}\\  
		\rotatebox{90}{\,\,\DAsetting{CS}{Rain}} & \rowqual{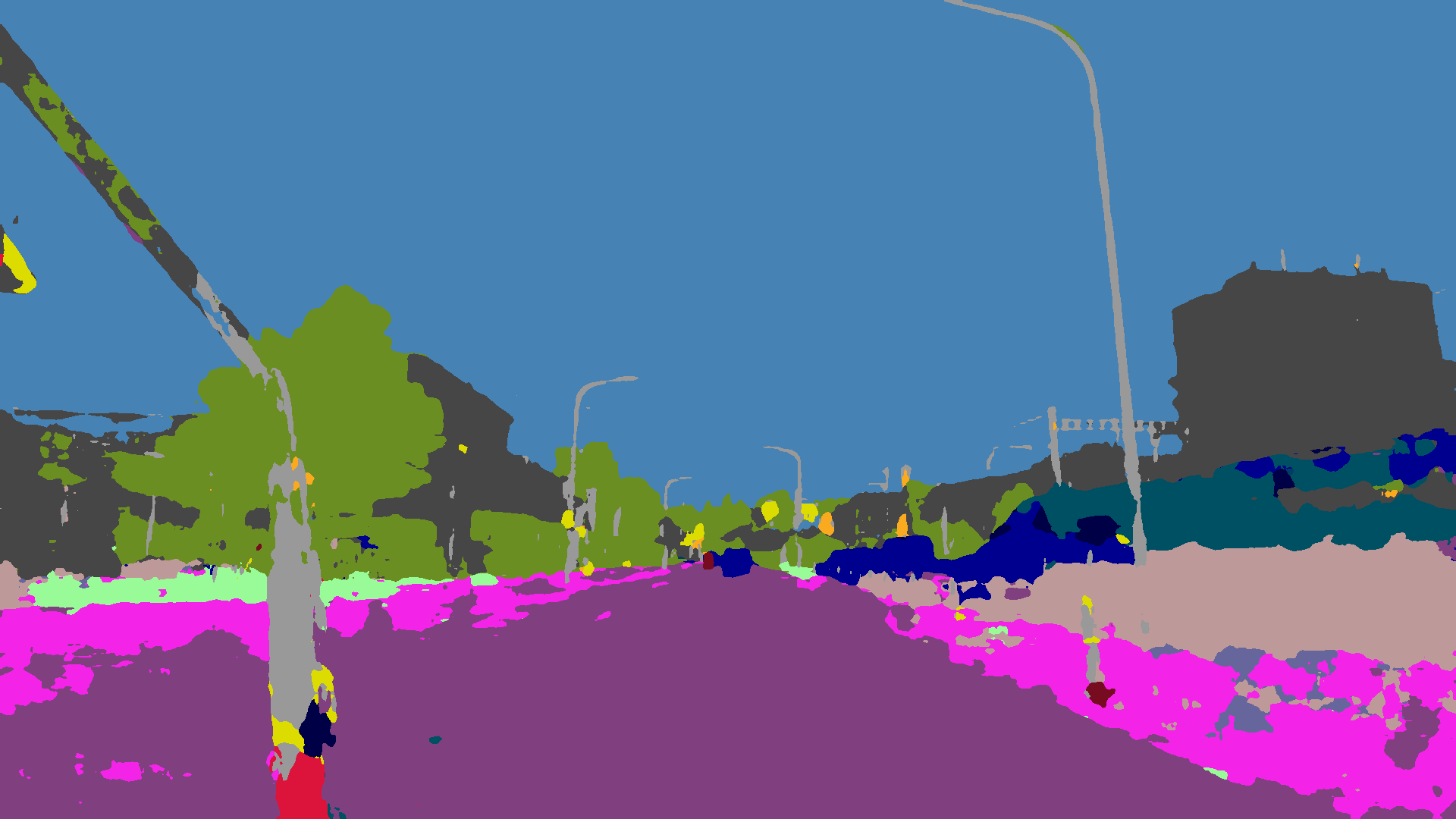}\\  
		\rotatebox{90}{\hspace{-0.3em}\DAsetting{GTA5}{Real}} & \rowqual{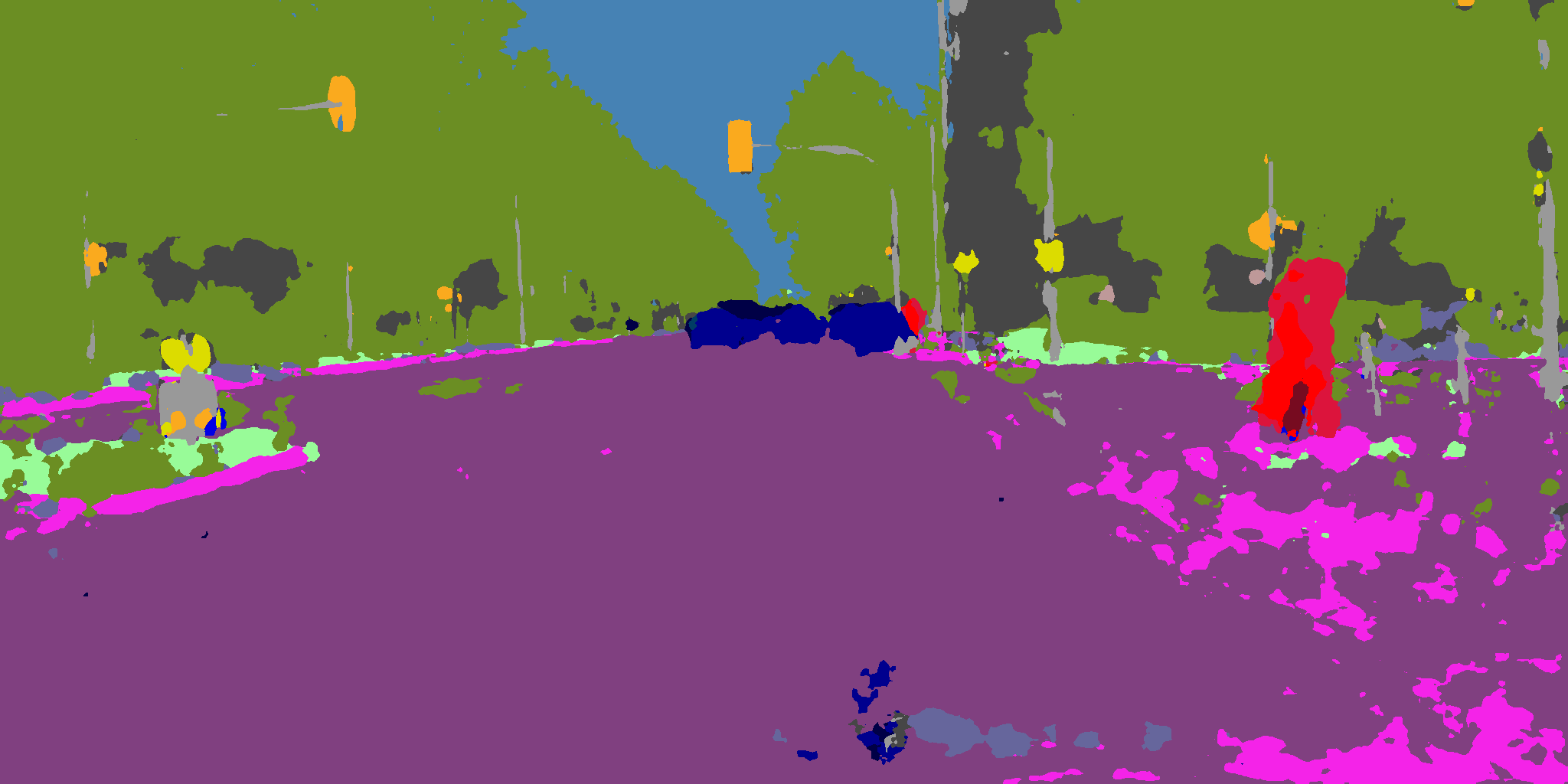}
	\end{tabular}
    }
    \smallskip\caption{\textbf{Qualitative results of zero-shot adaptation.} (\textit{Columns 1-2}) Input images and their ground truths; (\textit{Columns 3-4}) Segmentation results of source-only and ~\method models.}
    \label{fig:qualres}
\end{figure}

\DAsetting{Real}{synthetic} is an interesting though under-explored adaptation scenario. 
One potential application of \DAsetting{real}{synthetic} is for model validation in the industry, where some hazardous validations like driving accidents must be done in the virtual space.
Here we test if our zero-shot mechanism can be also applied to this particular setting.
Similarly, \method~outperforms both baselines.
Also in the reverse~\DAsetting{synthetic}{real} setting, again our method performs the best.
CLIPstyler undergoes almost $5\%$ drops in mIoU compared to source-only.

We argue on the simplicity of our method that only introduces minimal changes to the feature statistics, yet such changes are crucial for target adaptation.
CLIPstyler, designed for style transfer, involves training an additional StyleNet with $\approx{}615$k parameters for synthesizing the stylized images.
We base on the simplicity merit of~\method~to explain why it is more favorable than CLIPstyler for downstream tasks like semantic segmentation: the minimal statistics changes help avoiding significant drifts on the feature manifold which may otherwise result in unwanted errors.
For comparison, it takes us $0.3$ seconds to augment one source feature, while stylizing an image with CLIPstyler takes $65$ seconds (as measured on one RTX 2080TI GPU).

\begin{table}[t]

    		\setlength{\tabcolsep}{0.01\linewidth}
    \centering
  \resizebox{1.\linewidth}{!}{
  \begin{tabular}{clcc}
    \toprule
    Source & Target eval. & One-shot SM-PPM\,\cite{wu2022style} & Zero-shot \method \\
    \midrule
    \multirow{3}{*}{CS}
    & ACDC Night & 13.07\,/\,14.60~{\small($\Delta$=1.53)} & 18.31\,/\,\textbf{25.03} ~{\small({$\Delta$=6.72})}\\
    & ACDC Snow & 32.60\,/\,35.61 ~{\small({$\Delta$=3.01})} & 39.28\,/\,\textbf{43.90} ~{\small($\Delta$=4.62)}\\
    & ACDC Rain & 29.78\,/\,32.23 ~{\small($\Delta$=2.45)} & 38.20\,/\,\textbf{42.31} ~{\small($\Delta$=4.11)}\\
    \midrule
    GTA5 & CS & 30.48\,/\,39.32 ~{\small($\Delta$=8.84)} & 36.38\,/\,\textbf{40.08} ~{\small($\Delta$=3.70)} \\
    \bottomrule
  \end{tabular}}
  \smallskip\caption{\textbf{Comparison with SM-PPM.} 
  	Semantic segmentation performance (mIoU\%) for source\,/\,adapted models, and gain provided by adaptation ($\Delta$ in mIoU). For adaptation, SM-PPM (ResNet-101 DeepLabv2) has access to one target image, while \method~(ResNet-50 DeepLabv3+) leverages a target prompt and a text encoder.} 
  \label{tab:comparison_with_OSUDA}
\end{table}

We show in \cref{fig:qualres} qualitative examples of predictions from source-only and \method~models.
{We report class-wise performance in Supplementary Material.

\begin{table}
\setlength{\tabcolsep}{0.01\linewidth}
\renewcommand{\arraystretch}{1.2}
\centering
\newcolumntype{H}{>{\setbox0=\hbox\bgroup}c<{\egroup}@{}}
\newcommand{\MPrompt}[1]{\cellcolor{gray!34}``#1''}
\newcommand{\GPTRel}[1]{\multicolumn{1}{>{\columncolor{gray!34}}c}{\color{colrelprompt!100}\tabularCenterstack{c}{``#1''}}}
\newcommand{\GPTIrr}[1]{\multicolumn{4}{>{\columncolor{gray!34}}c}{\color{colirrprompt!100}\tabularCenterstack{c}{``#1''}}}
\newcommand{\best}[1]{\textbf{#1}}
\newcommand{\topfive}[1]{\underline{#1}}
\resizebox{1.0\linewidth}{!}{
    \begin{tabular}{ccccc} \toprule
	Method & ACDC Night & ACDC Snow & ACDC Rain & GTA5\quad\quad\quad{}\\ \midrule 
	\makecell{Source\\only} & 18.31\textcolor{white}{\vartn{0.00}} & 39.28\textcolor{white}{\vartn{0.00}} & 38.20\textcolor{white}{\vartn{0.00}} & 39.59\textcolor{white}{\vartn{0.00}}\\ \midrule
		\makecell{Trg} & \MPrompt{driving at night} & \MPrompt{driving in snow} & \MPrompt{driving under rain} & \MPrompt{driving in a game}\\
		& 25.03\vartn{0.48}  & 43.90\vartn{0.53} & 42.31\vartn{0.55} & 41.07\vartn{0.48}\\ \midrule
		\multirow{26}{*}{\rotatebox{90}{\hspace{-21.5em} {\color{colirrprompt!100}$\longleftarrow$Irrelevant}\quad\fbox{ChatGPT-generated}\quad {\color{colrelprompt!100}Relevant $\longrightarrow$}}} & 
            \GPTRel{operating a\\vehicle\\after sunset} & \GPTRel{operating a\\vehicle in\\snowy conditions} & \GPTRel{operating a\\vehicle in\\ wet conditions} & \GPTRel{piloting a\\vehicle in\\a virtual world}\\
		& 24.38\vartn{0.37} & 44.33\vartn{0.36} & 42.21\vartn{0.47} &  41.25\vartn{0.40}\\
		& \GPTRel{driving during\\the nighttime\\hours} & \GPTRel{driving on\\snow-covered\\roads} & \GPTRel{driving on\\rain-soaked\\roads} & \GPTRel{controlling a car\\ in a digital\\simulation} \\
		& \textbf{25.22}\vartn{0.64} & 43.56\vartn{0.62} & \textbf{42.51}\vartn{0.33} & 41.19\vartn{0.14} \\
		& \GPTRel{navigating\\the roads\\in darkness} & \GPTRel{piloting a\\vehicle in\\snowy terrain} & \GPTRel{navigating\\through rainfall\\while driving} & \GPTRel{maneuvering\\a vehicle in a\\ computerized racing\\ experience'} \\
		& 24.73\vartn{0.47}  & \textbf{44.67}\vartn{0.18} & 41.11\vartn{0.69} & 40.34\vartn{0.49}\\
		& \GPTRel{driving in\\low-light\\conditions} & \GPTRel{driving in\\wintry\\precipitation} & \GPTRel{driving in\\inclement\\weather} & \GPTRel{operating\\a transport \\in a video game\\environment} \\
		& 24.68\vartn{0.34}  & 43.11\vartn{0.56} & 40.68\vartn{0.37} & 41.34\vartn{0.42} \\
		& \GPTRel{travelling\\by car\\after dusk} & \GPTRel{travelling\\by car in\\a snowstorm} & \GPTRel{travelling by\\car during\\a downpour} & \GPTRel{navigating a\\machine through\\a digital\\driving simulation} \\
		& 24.89\vartn{0.24} & 43.83\vartn{0.17} & 42.05\vartn{0.35} & \textbf{41.86}\vartn{0.10} \\ \cmidrule(lr){2-2} \cmidrule(lr){3-3} \cmidrule(lr){4-4}
\cmidrule(lr){5-5}
		& \textit{24.82}\textcolor{white}{\vartn{0.00}} & \textit{43.90}\textcolor{white}{\vartn{0.00}} & \textit{41.81}\textcolor{white}{\vartn{0.00}} & \textit{41.18}\textcolor{white}{\vartn{0.00}} \\ \cmidrule{2-5}
            & \GPTIrr{mesmerizing northern lights display}\\
		&  20.05\vartn{0.77} &  40.07\vartn{0.66} & 38.43\vartn{0.82} & 37.98\vartn{0.31} \\
		& \GPTIrr{playful dolphins in the ocean}\\
		& 20.11\vartn{0.31} & 39.87\vartn{0.26} & 38.56\vartn{0.58} & 37.05\vartn{0.31}\\
		& \GPTIrr{breathtaking view from mountaintop}\\
		& 20.65\vartn{0.33} & 42.08\vartn{0.28} & 40.05\vartn{0.52} & 40.09\vartn{0.23}\\
		& \GPTIrr{cheerful sunflower field in bloom}\\
		& 21.10\vartn{0.50} & 39.85\vartn{0.68} & 40.09\vartn{0.41} & 37.93\vartn{0.55}\\
		& \GPTIrr{dramatic cliff overlooking the ocean}\\
		& 20.09\vartn{0.98} & 38.20\vartn{0.54} & 38.48\vartn{0.37} & 37.57\vartn{0.46}\\
		& \GPTIrr{majestic eagle in flight over mountains}\\
 		& 20.70\vartn{0.38} & 39.60\vartn{0.27} & 40.38\vartn{0.86} & 38.52\vartn{0.21}\\ 
            \cmidrule(lr){2-2} \cmidrule(lr){3-3} \cmidrule(lr){4-4}
\cmidrule(lr){5-5}
            & \textit{20.45}\textcolor{white}{\vartn{0.00}} & \textit{39.95}\textcolor{white}{\vartn{0.00}} & \textit{39.33}\textcolor{white}{\vartn{0.00}} & \textit{38.19}\textcolor{white}{\vartn{0.00}} \\ \bottomrule
    \end{tabular}}
    \smallskip\caption{\textbf{Effect of prompts on \method{}.} We show result for our $\prompt{}$ (top) as well as ChatGPT-generated \setlength{\fboxsep}{1pt}\colorbox{gray!34}{\relprompt{}} (middle) and \setlength{\fboxsep}{1pt}\colorbox{gray!34}{\irrprompt{}} (bottom). Please refer to \cref{sec:abla} for details. Best results (\textbf{bold}) are always obtained with 
    {\color{colrelprompt!100}relevant prompts} for which mean mIoU (\textit{italic}) also proves to be better.}
    \label{tab:text_effect_RN50_relevant}
    \label{tab:text_effect_RN50_irrelevant}
\end{table}

\smallskip\noindent\textbf{Comparison to one-shot UDA (OSUDA). }
We also compare \method~against SM-PPM \cite{wu2022style}\footnote{We use official code \url{https://github.com/W-zx-Y/SM-PPM}}, a state-of-the-art OSUDA method, 
see \cref{tab:comparison_with_OSUDA}. 
The OSUDA setting allows the access to a single unlabeled target domain image for DA. In SM-PPM, this image is considered as an anchor point for target style mining.
Using $5$ randomly  selected target images, we trained, with each one, five models with different random seeds. The reported mIoUs are averaged over the 25 resulting models.
We note that the absolute results of the two models are not directly comparable due to the differences in backbone~(ResNet-101 in SM-PPM \vs ResNet-50 in \method) and in segmentation framework~(DeepLabv2 in SM-PPM \vs DeepLabv3+ in \method).
We thus analyze the improvement of each method over the corresponding naive source-only baseline while taking into account the source-only performance.
We first notice that our source-only~(CLIP ResNet) performs better than \makebox{SM-PPM} source-only (ImageNet pretrained ResNet), demonstrating the overall robustness of the frozen CLIP-based model. 
In \DAsetting{Cityscapes}{
ACDC}, both absolute and relative improvements of \method~over source-only are greater than the ones of SM-PPM.
Overall, \method~exhibits on par or greater improvements over \makebox{SM-PPM}, despite the fact that our method is purely zero-shot.

\smallskip\noindent\textbf{{Qualitative results on {uncommon} conditions}.} Figure \ref{fig:long_tail_quali} shows some qualitative results, training on Cityscapes, and adapting to {uncommon} conditions {never found in datasets because they are either rare (\textit{sandstorm}), dangerous (\textit{fire}), or not labeled (\textit{old movie}). For all, \method{} improves over source-only,
which demonstrates its true benefit.}

\subsection{Ablation studies}
\label{sec:abla}

\noindent\textbf{$\prompt$ selection.~} 
Using any meaningful descriptions of the target domain, one should obtain similar adaptation gain with~\method{}.
{To verify this, we generate other \relprompt{}s by querying ChatGPT\footnote{OpenAI's chatbot \url{https://chat.openai.com/}} with \query{Give me 5 prompts that have the same exact meaning as [PROMPT]} using same prompts as in \cref{tab:main_results}.
Results in~\cref{tab:text_effect_RN50_relevant} show that 
adaptation gains are rather independent of the textual expression.
Inversely, we query \irrprompt{}s with \query{Give me 6 random prompts of length from 3 to 6 words describing a random photo}, which could result in negative transfer (See~\cref{tab:text_effect_RN50_irrelevant}).}  
By chance, small gains could occur; however we conjecture that such gains may originate from generalization by randomization rather than adaptation.

\begin{figure}[t!]
    \centering
    \resizebox{1.0\linewidth}{!}{%
        \setlength{\tabcolsep}{0.0015\linewidth}
        \small
        \begin{tabular}{ccc}
            \textbf{Input} & \textbf{Source-only} & \textbf{\method{}}\\
            \multicolumn{3}{c}{\cellcolor{gray!34}$\prompt$ = ``driving through fire''}\\
            \includegraphics[width=0.33\linewidth]{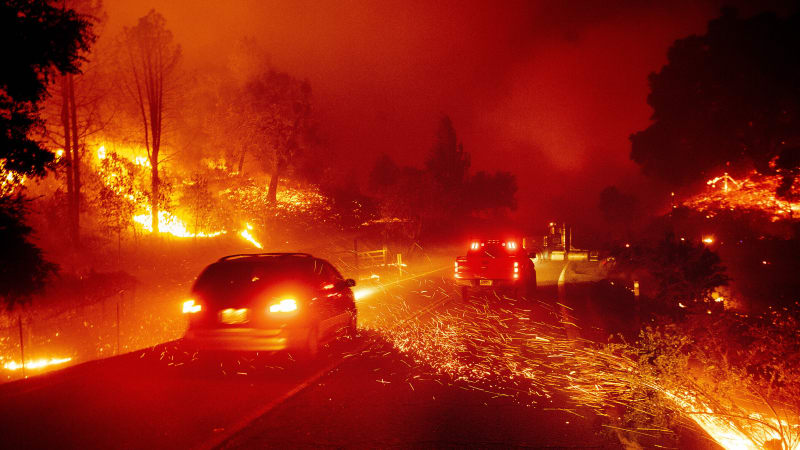} & \includegraphics[width=0.33\linewidth]{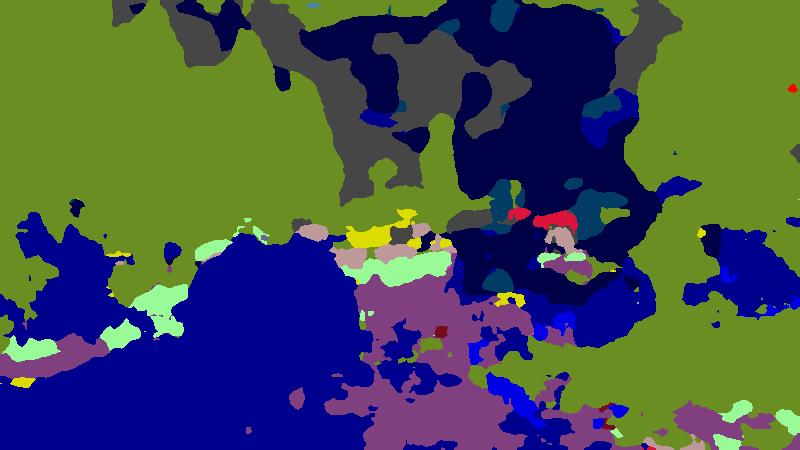} & \includegraphics[width=0.33\linewidth]{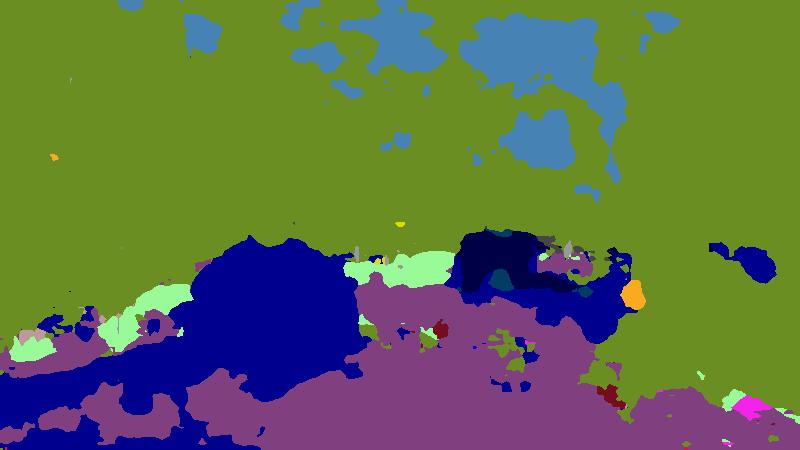}\\
            \includegraphics[width=0.33\linewidth,height=0.15\linewidth]{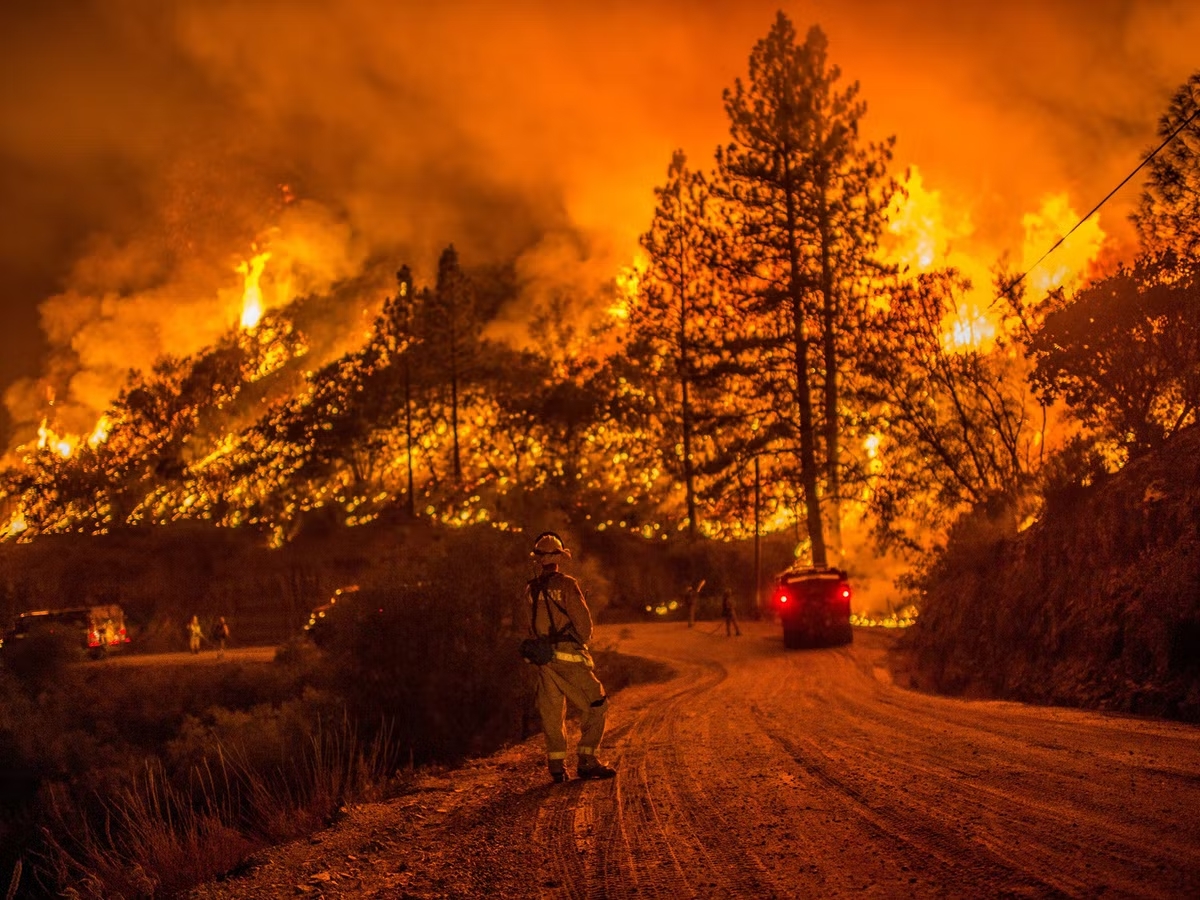} & \includegraphics[width=0.33\linewidth,height=0.15\linewidth]{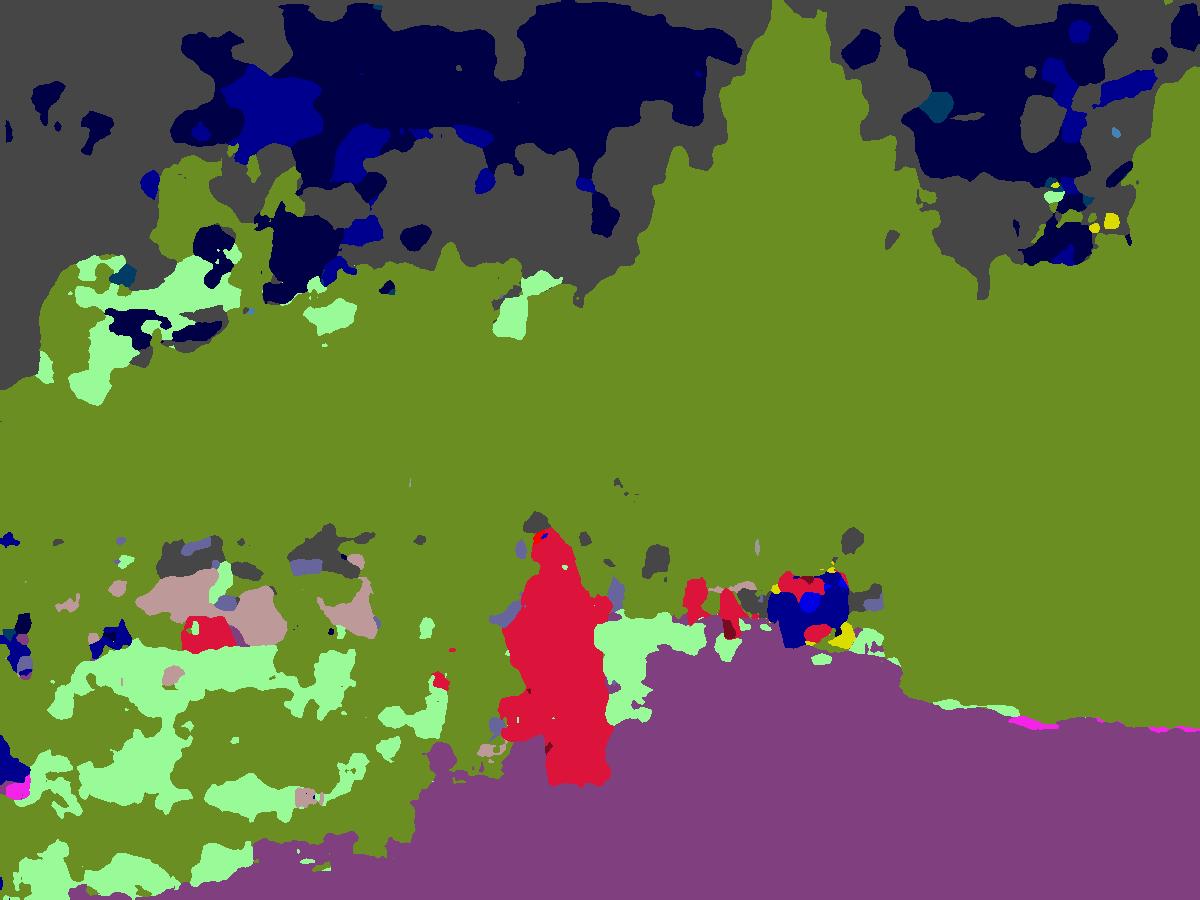} & \includegraphics[width=0.33\linewidth,height=0.15\linewidth]{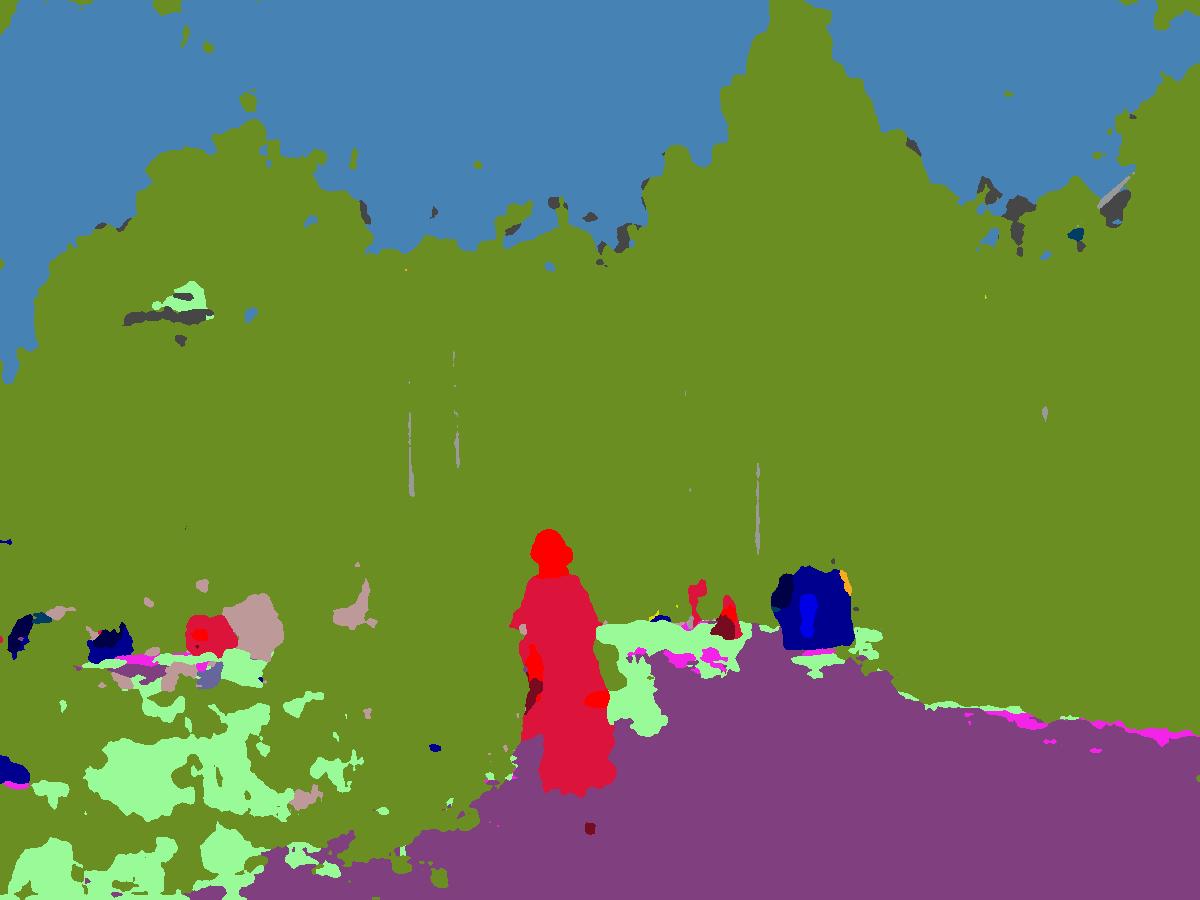}\\
            \multicolumn{3}{c}{~}\\
            \multicolumn{3}{c}{\cellcolor{gray!34}$\prompt$ = ``driving in sandstorm''}\\
            \includegraphics[width=0.33\linewidth,height=0.15\linewidth]{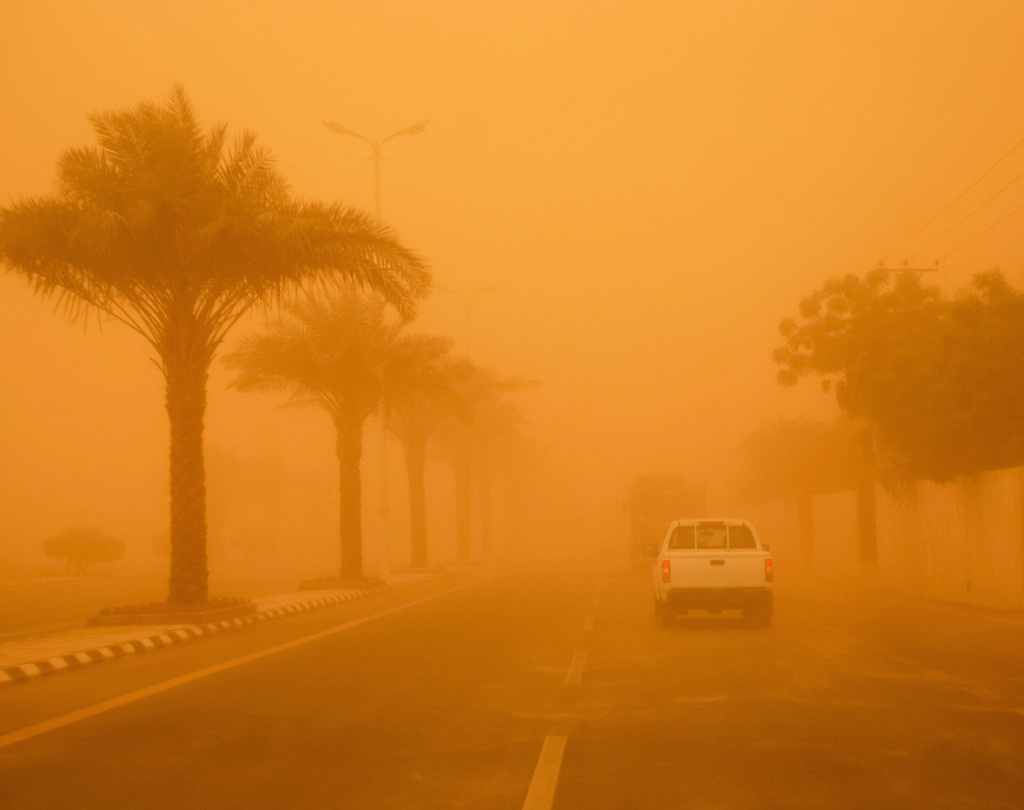} & \includegraphics[width=0.33\linewidth,height=0.15\linewidth]{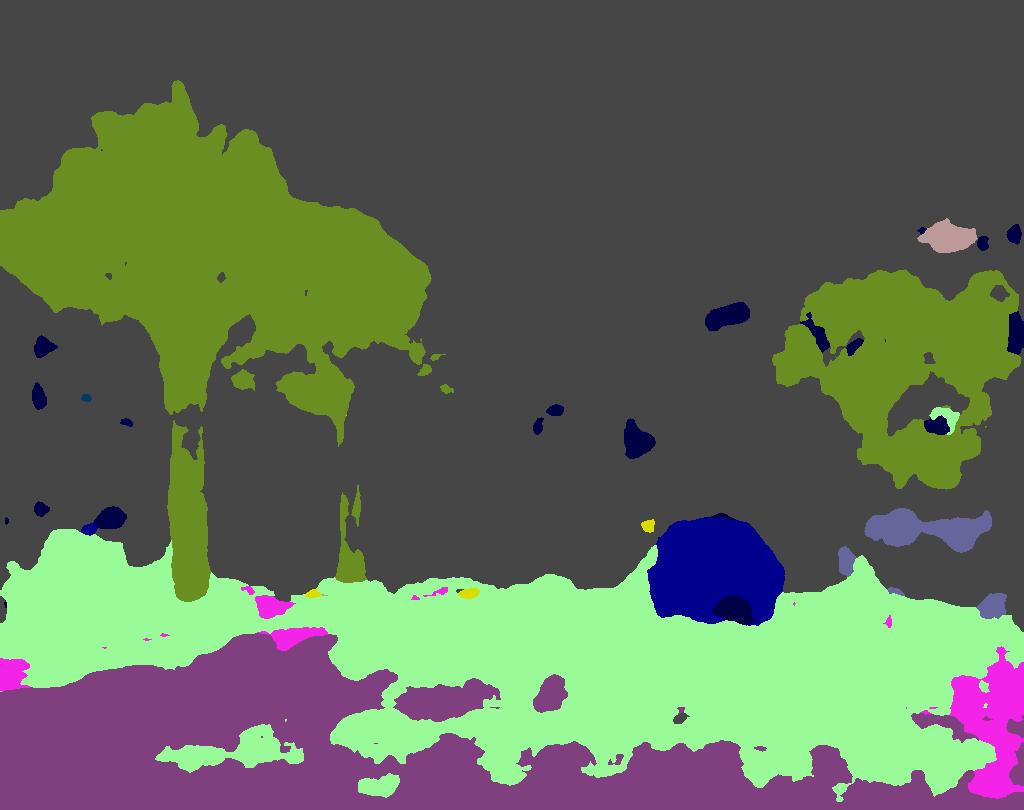} & \includegraphics[width=0.33\linewidth,height=0.15\linewidth]{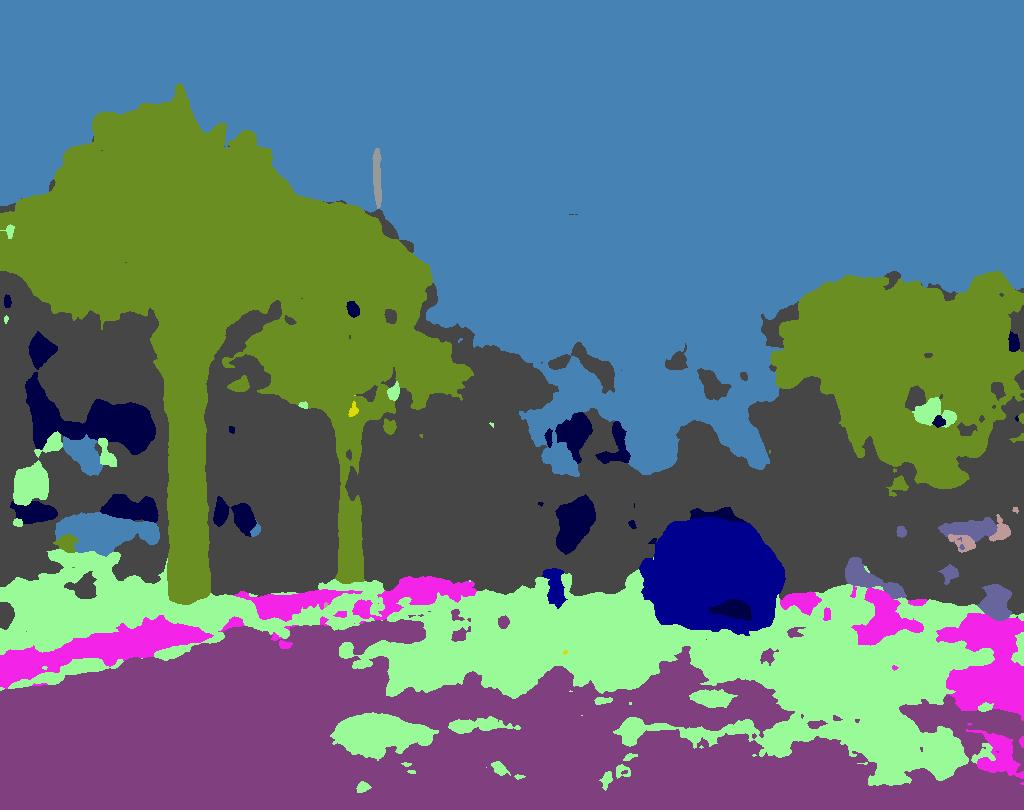}\\
            \includegraphics[width=0.33\linewidth,height=0.15\linewidth]{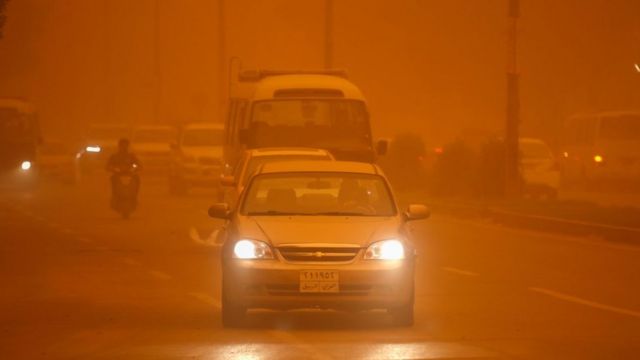} & \includegraphics[width=0.33\linewidth,height=0.15\linewidth]{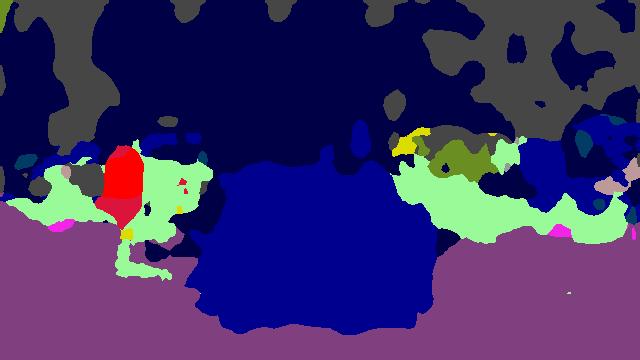} & \includegraphics[width=0.33\linewidth,height=0.15\linewidth]{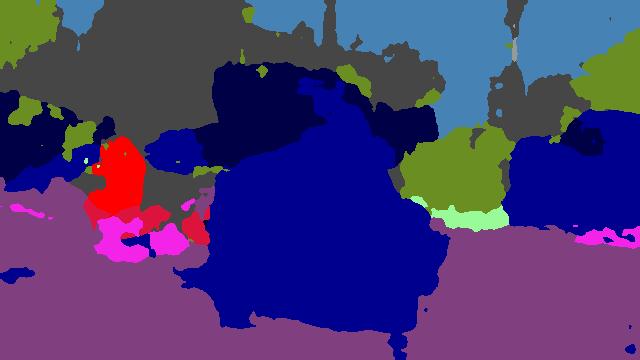}\\
            \multicolumn{3}{c}{~}\\
            \multicolumn{3}{c}{\cellcolor{gray!34}$\prompt$ = ``driving in old movie''}\\
            \includegraphics[width=0.33\linewidth,height=0.15\linewidth]{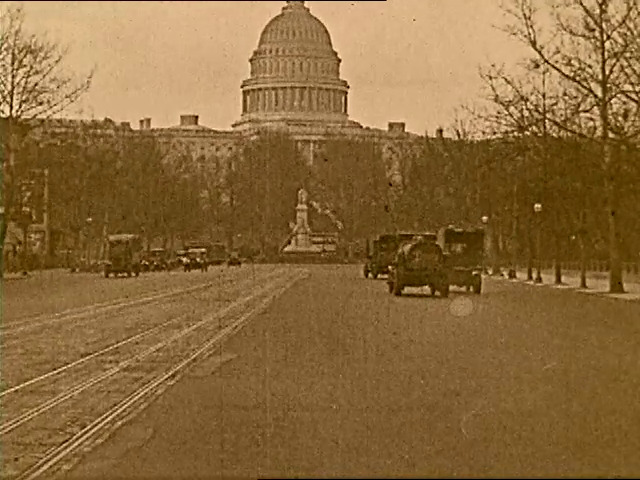} & \includegraphics[width=0.33\linewidth,height=0.15\linewidth]{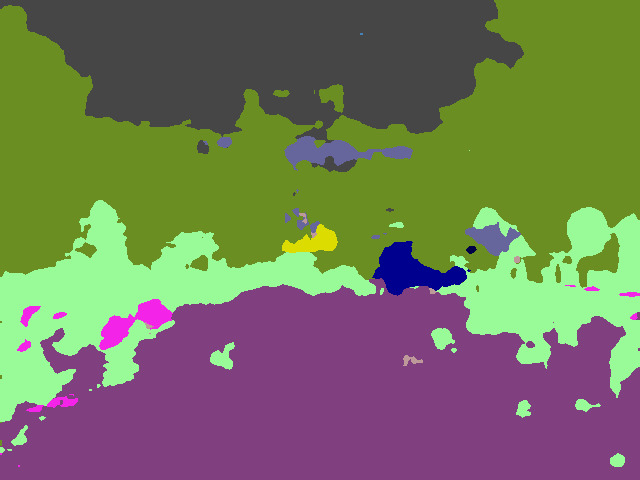} & \includegraphics[width=0.33\linewidth,height=0.15\linewidth]{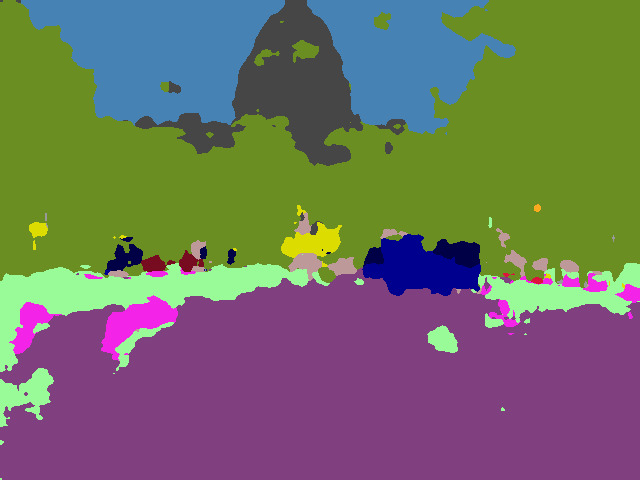}\\
            \includegraphics[width=0.33\linewidth,height=0.15\linewidth]{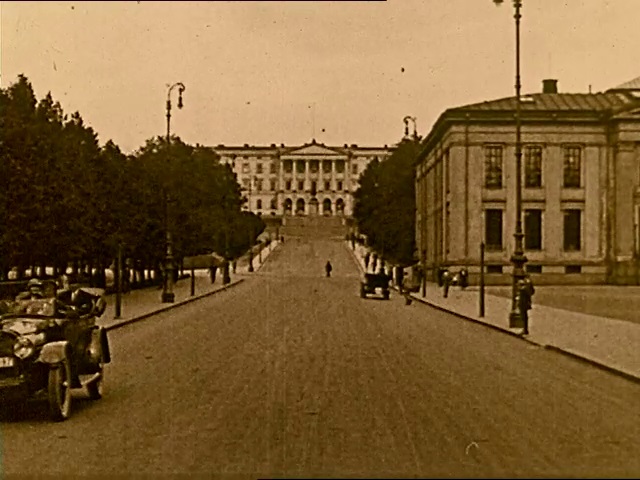} & \includegraphics[width=0.33\linewidth,height=0.15\linewidth]{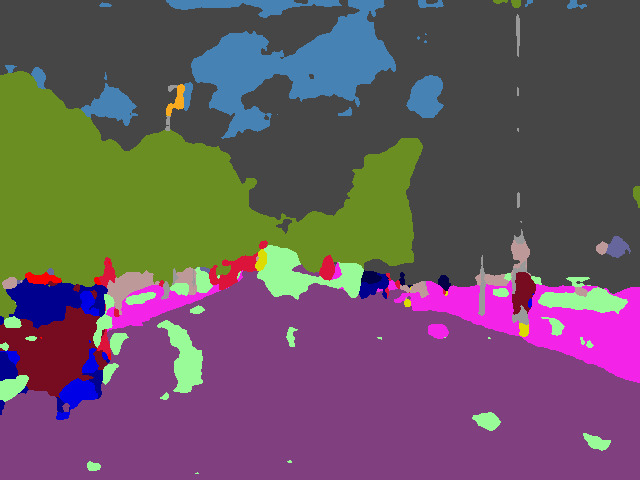} & \includegraphics[width=0.33\linewidth,height=0.15\linewidth]{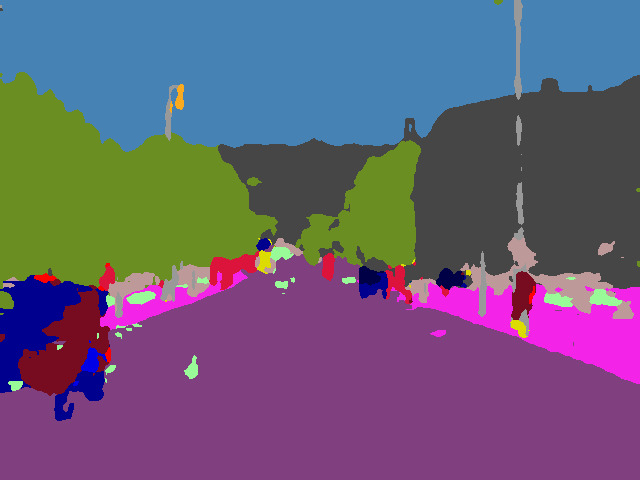}\\
        \end{tabular}
        \label{fig:long_tail}%
    }
    \smallskip
    \caption{\textbf{\method{} on {uncommon} conditions.} {Qualitative results here all use Cityscapes as source and \method{} uses {\setlength{\fboxsep}{1pt}\colorbox{gray!34}{$\prompt$}}.}}
    \label{fig:long_tail_quali}
\end{figure}

\smallskip\noindent\textbf{Choice of features to augment.}~
DeepLabV3+ 
segmenter takes as inputs both low-level features from \emph{Layer1} and high-level features from \emph{Layer4}.
In~\method, we only augment the \emph{Layer1} features and forward them through remaining layers 2-4 to obtain the \textit{Layer4} features. 
The input to the classifier is the concatenation of both.
We study in \cref{tab:layer_augment} if one should augment other features in addition to the ones in \emph{Layer1}: 
we observe 
the best performance with only \emph{Layer1} augmentation.
We conjecture that it is important to preserve the consistency between the two inputs to the classifier, \ie, \emph{Layer4} features should be derived from the augmented one from \emph{Layer1}.}

\smallskip\noindent\textbf{Number of mined styles.~}
Our experiments use always $|\stot{\mc{S}}| = |\src{\mc{D}}|$ but we study the effect of changing the number $|\stot{\mc{S}}|$ of styles on the target domain performance. By performing ablation on \DAsetting{CS}{Night} with $|\stot{\mc{S}}|=$ 1, 10, 100, 1000, 2975 (i.e., $|\src{\mc{D}}|$), we obtain 16.00\vartn{5.01}, 22.04\vartn{1.24}, 23.90\vartn{0.96}, 24.27\vartn{0.70}, 25.03\vartn{0.48} respectively. 
{For $|\stot{\mc{S}}| < |\src{\mc{D}}|$, the styles are sampled randomly from $\src{\mc{D}}$ and results are reported in average on 5 different samplings. Interestingly, we observe that the variance decreases with the increase of $|\stot{\mc{S}}|$. Results also suggest that only few styles (\eg $|\stot{\mc{S}}|= 10$) could be sufficient for feature translation, similarly to few-shot image-to-image translation~\cite{pizzati2022manifest}, though at the cost of higher variance.
}

\smallskip\noindent\textbf{Partial unfreezing of the backbone.~}
While our experiments use a frozen backbone due to the observed good out-of-distribution performance (\cref{tab:baselines}), we highlight that during training only \emph{Layer1 must} be frozen to preserve its activation space where augmentations are done; the remaining three layers could be optionally fine-tuned. Results in ~\cref{tab:p_unfrozen} show that freezing the whole backbone (\ie, \emph{Layer1-4}) achieves the best results.
In all cases, \method consistently improves the performance over source-only.

\begin{table}[t]

  \setlength{\tabcolsep}{0.01\linewidth}
  \centering
	
  \begin{tabular}{ccccc}
    \toprule
    \textit{Layer1} & \textit{Layer2} & \textit{Layer3} & \textit{Layer4} &  ACDC Night \\
    \midrule
    \checkmark & \xmark & \xmark & \xmark & \textbf{25.03}\vartn{0.48} \\
    \checkmark & \checkmark & \xmark & \xmark & 23.43\vartn{0.51} \\
    \checkmark & \xmark & \checkmark & \xmark & 22.93\vartn{0.53} \\
    \checkmark & \xmark & \xmark & \checkmark & 21.05\vartn{0.55} \\
    \bottomrule
  \end{tabular}

   \smallskip\caption{{\textbf{Impact of selected layers for augmentation}. Performance (mIoU) of \method{}'s \DAsetting{day}{night} adaptation for different choices of ResNet layers for feature augmentation. In addition to augmenting features of \textit{Layer1} ({Row\,1)}, one can augment \textit{Layer2} or \textit{Layer3} features ({Rows\,2-3}), or of \textit{Layer4} directly ({Row\,4}).}
   } 
 \label{tab:layer_augment}
\end{table}

\begin{table}[!t]
\setlength{\tabcolsep}{0.01\linewidth}
\centering
 \resizebox{1.0\linewidth}{!}{
	\begin{tabular}{lllll}
		\toprule
		  Method & Night & Snow & Rain & GTA5\\
		\midrule
        src-only* &  18.31\textcolor{white}{\vartn{0.00}} & 39.28\textcolor{white}{\vartn{0.00}} & 38.20\textcolor{white}{\vartn{0.00}} & 39.59\textcolor{white}{\vartn{0.00}} \\
        \method{}* & \textbf{25.03}\vartn{0.48} & \textbf{43.90}\vartn{0.53} & \textbf{42.31}\vartn{0.55} & \textbf{41.07}\vartn{0.48} \\
        \midrule
        src-only (\emph{Layer1} \texttwemoji{2744}) &  9.60\textcolor{white}{\vartn{0.00}} & 30.99\textcolor{white}{\vartn{0.00}} & 30.89\textcolor{white}{\vartn{0,00}} & 29.38\textcolor{white}{\vartn{0.00}} \\
        \method{} (\emph{Layer1} \texttwemoji{2744}) & 19.43\vartn{0.69} & 37.80\vartn{2.65} & 40.71\vartn{1.06} & 39.09\vartn{1.23}\\
        \bottomrule
	\end{tabular}
 }
	\smallskip\caption{\footnotesize{\textbf{\method{} when freezing (\texttwemoji{2744}) only \emph{Layer1}}.
 Both models with *, reported in \cref{tab:main_results},
 freeze the whole backbone \emph{Layer1-4}.}}
    \label{tab:p_unfrozen}
\end{table}

\begin{table}[t]
    		\setlength{\tabcolsep}{0.01\linewidth}
    \centering
	\resizebox{1.0\linewidth}{!}{
  \begin{tabular}{lllll}
    \toprule
    Method & Night & Snow & Rain & GTA5\\
    \midrule
    Source-only & 18.31\textcolor{white}{\vartn{0.00}} & 39.28\textcolor{white}{\vartn{0.00}} & 38.20\textcolor{white}{\vartn{0.00}} & 39.59\textcolor{white}{\vartn{0.00}}\\ 
    Source-only-G & 21.07\textcolor{white}{\vartn{0.00}} & 42.84\textcolor{white}{\vartn{0.00}} & 42.38\textcolor{white}{\vartn{0.00}} & 41.54\textcolor{white}{\vartn{0.00}}\\
    \method{}-G & \textbf{24.86}\vartn{0.70} & 44.34\vartn{0.36} & 43.17\vartn{0.63} & 41.73\vartn{0.39}\\
    \method{}-G+style-mix & 24.18\vartn{0.23} & \textbf{44.46}\vartn{0.34} & \textbf{43.56}\vartn{0.46} & \textbf{42.98}\vartn{0.12} \\
    \bottomrule
  \end{tabular}
  }
  \smallskip\caption{\textbf{Generalization with \method{}.} Source-only-G model is enhanced with a domain generalization technique. Training~\method{} from Source-only-G (`\method{}-G') brings improvements. \makebox{`style-mix'}: style mixing as in \cite{wu2022style}.
  }
  \label{tab:generalization}
\end{table}

\subsection{Further discussion}
\label{sec:discussion}

\noindent\textbf{Generalization with \method{}.~}
Inspired by the observation that some unrelated prompts improve performance on target domains (see \cref{tab:text_effect_RN50_irrelevant}), we study how \method{} can benefit from general style augmentation.
First, we coin \makebox{``Source-only-G''} the generalized source-only model where 
we augment features by shifting the per-channel $(\boldsymbol{\mu},\boldsymbol{\sigma})$ with Gaussian noises sampled for each batch of features, such that the signal to noise ratio is 20 dB. 
{This source-only variant takes inspiration {from~\cite{fan2023towards}} where simple perturbations of feature channel statistics could help achieve SOTA generalization performance in object detection.}
\cref{tab:generalization} shows that {Source-only-G} always improves over Source-only, demonstrating a generalization capability.
When applying our zero-shot adaptation on Source-only-G (denoted~\makebox{``\method{}-G''}), target performance again improves -- always performing best on the desired target. 
{Performance is further boosted by the style mixing strategy used in~\cite{wu2022style}, \ie source and augmented features statistics being linearly mixed.}

\smallskip\noindent\textbf{Effect of priors.~}
We now discuss existing techniques that approach zero-shot DA with different priors, revealing the potential combinations of different orthogonal methods. In \cref{tab:effectpriors}, we report zero-shot results of CIConv~\cite{lengyel2021zero} using physics priors, compared against CLIPstyler and~\method, which use textual priors, \ie, prompts.
We also include the one-shot SM-PPM~\cite{wu2022style} model 
as the single target sample it requires can be considered as a prior.
CIConv, a dedicated physics-inspired layer, is proven effective in enhancing backbone robustness on night scenes.
The layer could be straightforwardly included in the CLIP image encoder to achieve the same effect.
Albeit interesting, this combination would however require extremely high computational resources to re-train a CLIP variant equipped with CIConv.
We leave open such a combination, as well as others like (i) combining image-level (CLIPstyler) and feature-level augmentation (\method)  or (ii) additionally using style information from one target sample (like in SM-PPM) to help in guiding better the feature augmentation.

\smallskip\noindent\textbf{Other architectures.~}  We show in \cref{tab:fpn} consistent gains brought by \method{} using new backbone (RN101~\cite{he2016deep}) and segmenter (semantic FPN~\cite{kirillov2019panoptic}).

\begin{table}[t]
    \setlength{\tabcolsep}{0.01\linewidth}
    \centering
    \begin{tabular}{lll}
    	\toprule
    	Method & Prior & \quad\quad ACDC Night\\
    	\midrule
    	CIConv*~\cite{lengyel2021zero} & physics & 30.60\,/\,34.50 {\small ($\Delta$=3.90)} \\ 
    	SM-PPM~\cite{wu2022style} & 1 target image & 13.07\,/\,14.60 \small{($\Delta$=1.53)}\\
     	CLIPstyler~\cite{kwon2022clipstyler} & 1 prompt & 18.31\,/\,21.38 {\small($\Delta$=3.07)}\\
    	\method & 1 prompt & 18.31\,/\,25.03 \small{($\Delta$=6.72)}\\
    	\bottomrule
    \end{tabular}\\
    \smallskip{\scriptsize * Results of CIConv are on DarkZurich, a subset of ACDC Night~\cite{sakaridis2021acdc}.}
\smallskip\caption{\textbf{Effect of different priors for zero-shot/one-shot adaptation.} We report mIoU\% for source-only / adapted models, and gain brought by adaptation ($\Delta$ in mIoU). Note that \cite{lengyel2021zero,wu2022style} use a deeper backbone making results not directly comparable.}
\label{tab:effectpriors}
\end{table}

\begin{table}[!t]
\setlength{\tabcolsep}{0.01\linewidth}
\centering
\resizebox{1.0\linewidth}{!}{
\begin{tabular}{llllll}
    \toprule
    Backbone & Method & Night & Snow & Rain & GTA5\\
    \midrule
    \multirow{2}{*}{Sem.\,FPN} & src-only &  18.10\textcolor{white}{\vartn{0.00}} & 35.75\textcolor{white}{\vartn{0.00}} & 36.07\textcolor{white}{\vartn{0.00}} & 40.67\textcolor{white}{\vartn{0.00}} \\
    & \method& \textbf{21.48}\vartn{0.15} & \textbf{39.55}\vartn{0.13} & \textbf{38.34}\vartn{0.29} & \textbf{41.59}\vartn{0.24} \\
    \midrule
    \multirow{2}{*}{DLv3+} & src-only & 22.17\textcolor{white}{\vartn{0.00}} & 44.53\textcolor{white}{\vartn{0.00}} & 42.53\textcolor{white}{\vartn{0.00}} & 40.49\textcolor{white}{\vartn{0.00}} \\
    & \method & \textbf{26.54}\vartn{0.12} & \textbf{46.71}\vartn{0.43} & \textbf{46.36}\vartn{0.20} & \textbf{43.17}\vartn{0.13} \\
    \bottomrule 
\end{tabular}
}
\smallskip\caption{\textbf{\method{} with different architectures.} Backbones are RN50 for Semantic FPN (`Sem.\,FPN') and RN101 for DeepLabV3+ (`DLv3+').}
\label{tab:fpn}
\end{table}

\section{\method{} for other tasks}
\label{sec:extensiontasks}
\method{} operates at the features level, which makes it task-agnostic. We show in the following the effectiveness of our method for object detection and image classification.

\smallskip\noindent\textbf{\method{} for Object Detection.~}
We report in~\cref{tab:poda_od_results} some 
results when straightforwardly applying~\method{} to object detection.
Our Faster-RCNN~\cite{ren2015faster} models, initialized with two backbones, are trained on two source datasets, either Cityscapes or the Day-Clear split in Diverse Weather Dataset (DWD)~\cite{wu2022single}.
We report adaptation results on Cityscapes-Foggy~\cite{sakaridis2018semantic} and four other conditions in DWD.
For zero-shot feature augmentation in~\method{}, we use simple prompts and take the default optimization parameters in previous experiments.
~\method{} obtains on par or better results than
UDA methods~\cite{chen2021scale,rezaeianaran2021seeking} {(which use target images)} and domain generalization methods~\cite{fan2023towards,wu2022single,vidit2023clip}. We also experimented with YOLOF~\cite{chen2021you} for object detection in CS$\shortrightarrow$Foggy;~\method{} reaches $35.4\%$, improving $1.5\%$ from the source-only model.
These 
results open up potential combinations of~\method{} with generalization techniques like~\cite{fan2023towards} and~\cite{vidit2023clip} for object detection. 

\begin{table}[t!]
    \setlength{\tabcolsep}{0.01\linewidth}
    \centering
    \resizebox{1.0\linewidth}{!}{
    \begin{tabular}{lcccccc}
        \toprule
        &&\multirow{3}{*}{\makecell{CS$\shortrightarrow$ CS\\Foggy}} & \multicolumn{4}{c}{DWD-Day Clear $\shortrightarrow$} \\ \cmidrule(lr){4-7}
        Method & Target &  &\makecell{Night\\Clear} & \makecell{Dusk\\Rainy} & \parbox{0.8cm}{Night\\Rainy}&\parbox{0.8cm}{Day\\Foggy}  \\
         \midrule
         DA-Faster~\cite{chen2021scale} & \checkmark & 32.0&-&-&-&-\\
         ViSGA~\cite{rezaeianaran2021seeking} & \checkmark & 43.3&-&-&-&-\\
         NP+~\cite{fan2023towards} & \xmark & 46.3&-&-&-&-\\
         S-DGOD~\cite{wu2022single} & \xmark & -&36.6&28.2&16.6&33.5\\
         CLIP The Gap~\cite{vidit2023clip} & \xmark & -&36.9&32.3&18.7&38.5\\
         \method{} & \xmark & \textbf{47.3}& \textbf{43.4} & \textbf{40.2} & \textbf{20.5} & \textbf{44.4} \\
         \bottomrule
    \end{tabular}
    }
    \smallskip\caption{\textbf{\method for object detection (mAP\%)}. For Cityscapes$\shortrightarrow$Cityscapes-Foggy adaptation, the backbone is ResNet-50, while it is ResNet-101 for adaption from DWD-Day-Clear to other conditions in DWD.}
    \label{tab:poda_od_results}
\end{table}

\smallskip\noindent \textbf{\method{} for Image Classification.}
We show that \method{} can be also applied for image classification. We use the same augmentation strategy to adapt a linear probe on top of CLIP-RN50 features.
In a first experiment we train a linear classifier on the features of CUB-200 dataset~\cite{wah2011caltech} of 200 real bird species; we then perform zero-shot adaptation to classify bird paintings of CUB-200-Paintings dataset~\cite{wang2020progressive} using the single prompt \makebox{``Painting of a bird''}. In our second experiment, we address the color bias in Colored MNIST~\cite{arjovsky2019invariant}; while for training, \textcolor{red}{even} and \textcolor{blue}{odd} digits are colored \textcolor{red}{red} and \textcolor{blue}{blue} respectively, the test digits are randomly colored. We augment training digit features using the \makebox{``Blue digit''} and \makebox{``Red digit''} prompts for \textcolor{red}{even} and \textcolor{blue}{odd} digits respectively, and create a separate set for each one to prevent styles from leaking, \ie to avoid trivially using ``red'' styles coming from \textcolor{red}{even} digits to augment \textcolor{blue}{odd} digits features and vice versa. 
Results in~\cref{tab:classification} show that \method significantly improves over the source-only models. 

\begin{table}[t]
    \setlength{\tabcolsep}{0.01\linewidth}
    \centering
    \begin{tabular}{lll}
    	\toprule
    	\makecell{Method} & \makecell{CUB-200 \\ paintings} & \makecell{Colored \\ MNIST}\\
    	\midrule
    	src-only & 28.90\textcolor{white}{\vartn{0.00}} & 55.83 \textcolor{white}{\vartn{0.00}}\\ 
    	\method &  \textbf{30.91}\vartn{0.69} & \textbf{64.16}\vartn{0.41}\\
    	\bottomrule
    \end{tabular}
\smallskip
\caption{\textbf{\method for image classification (acc\%).} The backbone is ResNet-50, and a linear classifier is fit on top of the features. The source domains are CUB-200 (\textit{real} bird images) and colored MNIST with color bias, in second and third columns respectively.}
\label{tab:classification}
\end{table}

\section{Conclusion}
In this work, we leverage the powerful zero-shot ability of the CLIP model to make possible a new challenging task of domain adaptation using prompts.
We propose a cost-effective feature augmentation mechanism that adjusts the style-specific statistics of source features to synthesize augmented features in the target domain, guided by domain prompts in natural language.
Extensive experiments have proven the effectiveness of our framework for semantic segmentation in particular. They also show its applicability to other tasks and various backbones. 
Our line of research aligns with the collective efforts of the community to leverage large-scale pre-trained models (so-called ``foundation models''~\cite{bommasani2021opportunities}) for 
data- and label-efficient training of perception models for real-world applications.

\smallskip\noindent
\textbf{Acknowledgment.~} This work was partially funded by French project SIGHT (ANR-20-CE23-0016). 
The authors also thank Ivan Lopes and Fabio Pizzati for their kind proofreading.

\appendix

\section{Overall pseudo-code of \method{}}
\cref{algo:PODA_overall} presents the 
high-level pseudo-code of \method{}: from \emph{source-only} training as model initialization, to prompt-driven feature augmentation, to zero-shot model adaptation.

\section{Experimental details}

\noindent\textbf{Feature augmentation.} 
PIN operates on image features.
For augmentation, we optimize ($\bs{\mu},\bs{\sigma}$) of source feature map $\src{\mb{f}}$; it is done in batches for the sake of speed.
We fix the batch size $b=16$ and the learning rate $lr=1.0$.

\smallskip\noindent\textbf{Style mixing.} 
In the 
discussion of \method (\cref{sec:discussion} and \cref{tab:generalization}), 
we presented the performance gains that style-mixing~\cite{wu2022style} brings to our method in three settings. 
By randomly mixing original and augmented statistics, we introduce certain perturbations to the final augmented features.
The mixed statistics $\bs{\mu}_{\text{mix}},\bs{\sigma}_{\text{mix}}$ are given by:
\begin{align}
	\bs{\mu}_{\text{mix}} = \bs{\alpha}\trg{\bs{\mu}} + (1-\bs{\alpha})\src{\bs{\mu}}\,, \\
	\bs{\sigma}_{\text{mix}} = \bs{\alpha}\trg{\bs{\sigma}} + (1-\bs{\alpha})\src{\bs{\sigma}}\,, 
\end{align}
where $\bs{\alpha} \in \mathbb{R}^{c}$ are per-channel mixing weights uniformly sampled in $[0,1]$, similarly
to~\cite{wu2022style}; multiplications are element-wise. Finally, the augmented features are computed as follows:
\begin{align}
	\mb{f}_{\text{s}\shortrightarrow\text{t}} = \texttt{PIN}(\src{\mb{f}}, \bs{\mu}_{\text{mix}}, \bs{\sigma}_{\text{mix}}), 
\end{align}
with prompt-driven instance normalization $\texttt{PIN}$ defined in \cref{eqn:PIN}.

\section{Additional experiments}

\noindent\textbf{Effect of style mining initialization.}
In our feature optimization step, we initialize $(\bs{\mu},\bs{\sigma})$ with $(\mu(\src{\mb{f}}),\sigma(\src{\mb{f}}))$.
In~\cref{tab:init_effect}, we report results using different initialization strategies. 
Starting from pre-defined or random initialization, instead of from original statistics, degrades badly the performance.
As we do not use any regularization term in the CLIP cosine distance loss, we argue that initializing the optimized statistics with those of the source images is a form of regularization, favoring augmented features in a neighborhood of $\src{\emb{\mb{f}}}$ and better preserving the semantics.

\begin{algorithm}[t!]
	\small
	\SetAlgoLined
	\SetKwFunction{Train}{train}
	\SetKwFunction{Fn}{fine-tune}
	\SetKwFunction{Feat}{feat-ext}
	\SetKwFunction{Init}{init}
	\SetKwFunction{Mine}{style-mining}
	\SetKwInOut{Input}{Input}  
	\Input{Source dataset $\src{\mc{D}}=\{(\src{\mb{x}}, \src{\mb{y}})\}$ \\
		CLIP encoders $\EncI$ 
		and $\EncT$
		\\
		Target domain description $\prompt$ \\
		Feature backbone $M_\text{feat} \xleftarrow{} \EncI$ \\
		Source model: $M = (M_\text{feat}, M_\text{cls})$\\
	}
	\KwResult{Target-adapted model $M' = (M_\text{feat}, M'_\text{cls})$}
	\tcp{Initialization}
	$\promptFeat = \EncT(\prompt)$ \\
	$M_\text{cls} \gets \Train(M_\text{cls}, \src{\mc{D}})$ \Comment{\textcolor{codeblue}{source-only training} } \\
	\tcp{Feature Augmentation}
	$\src{\mc{F}} \gets \Feat(M_\text{feat}, \{\src{\mb{x}}\})$  \\
	$\stot{\mc{S}} \gets \texttt{augment}(\src{\mc{F}}, \promptFeat)$  \\
	\tcp{Adaptation}
	$M'_\text{cls} \gets \Fn(M_\text{cls}, \src{\mc{F}}, \stot{\mc{S}}, \{\src{\mb{y}}\})$ \makebox{\Comment{\textcolor{codeblue}{fine-tuning}}} \\
	\caption{Prompt-driven Zero-shot DA}
	\label{algo:PODA_overall}
\end{algorithm}

\begin{table}[t]
	\setlength{\tabcolsep}{0.015\linewidth}
	\centering
	\begin{tabular}{ccr}
		\toprule
		$\bs{\mu}^{0}$ & $\bs{\sigma}^{0}$ & mIoU\quad\quad \\
		\midrule
		$\mu(\src{\mb{f}})$ & $\sigma(\src{\mb{f}})$ & \textbf{25.03}\vartn{0.48} \\
        $\mathbf{0}$ & $\mathbf{1}$ & 8.59\vartn{0.82} \\
	    $\sim \mathcal{N}(\mathbf{0},\mathbf{I})$ & $\sim \mathcal{N}(\mathbf{0},\mathbf{I})$  & 6.80\vartn{0.92} \\
		\bottomrule
	\end{tabular}
 
	\smallskip\caption{\textbf{Effect of style initialization.} Performance (in mIoU) of \method{} on ACDC-Night val set (Cityscapes as source), with different style statistics initializations. Starting from source images' statistics works substantially better.}
\label{tab:init_effect}
\end{table}

\begin{figure}[t]
     \centering
     \includegraphics[width=\linewidth]{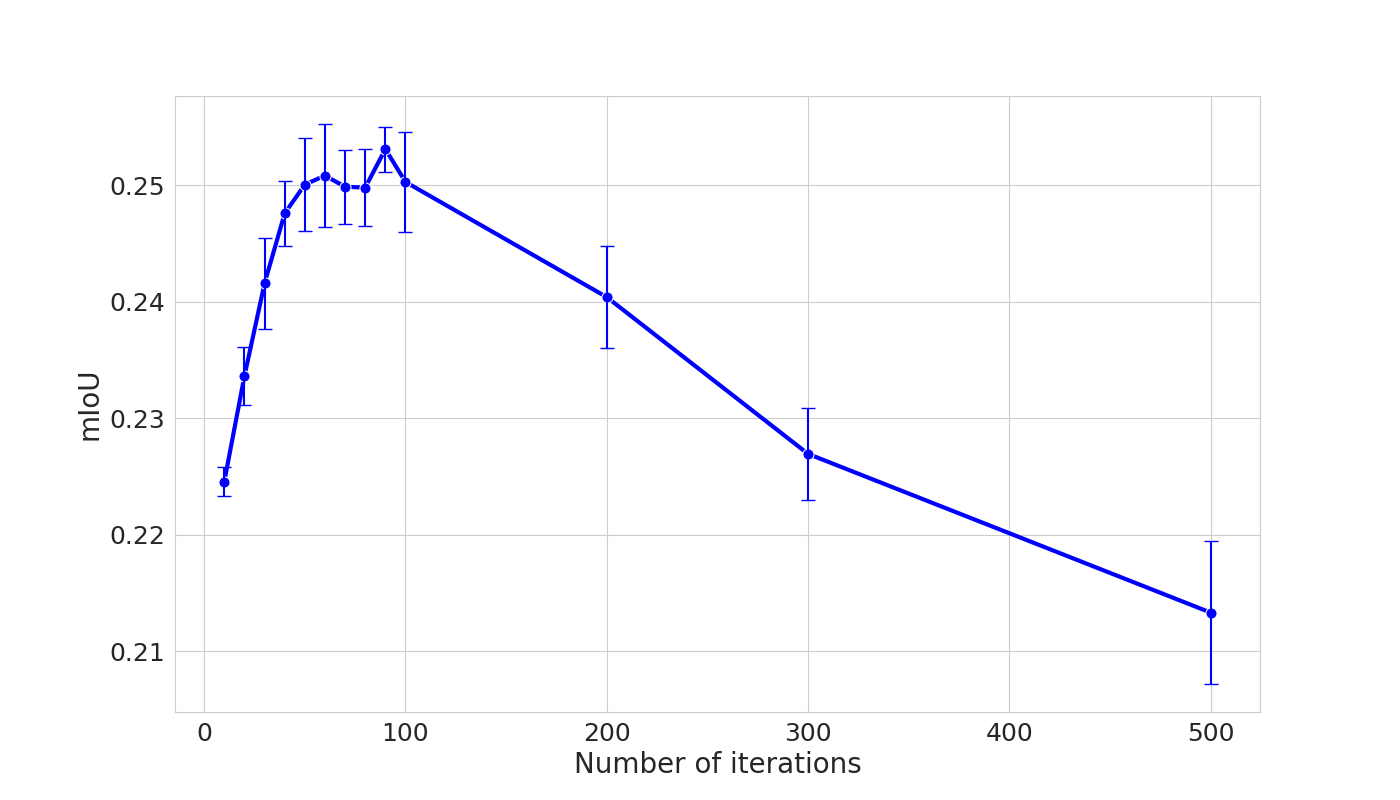}
     \caption{\textbf{Effect of the number of optimization iterations.} Performance (mIoU \%) of \method{} adaptation from Cityscapes to ACDC-Night as a function of the number of statistics optimization iterations. The values are averages over 5 runs and the bars represent the standard deviation.}
    \label{fig:ablation_iters}
\end{figure}

\smallskip\noindent\textbf{Optimization steps.}
In all our experiments, $100$ iterations of optimization are performed for each batch of source features. We show in~\cref{fig:ablation_iters} the effect of the total number of iterations. We see an inflection point at around 80-100 iterations. Using   
few iterations is not sufficient for style alignment. Above $100$, we also observe a performance drop.
We refer to~\cite{kwon2022clipstyler} and argue for the ``over-stylization'' problem in this case.

\begin{figure}[t!]
     \centering
         \includegraphics[width=\linewidth]{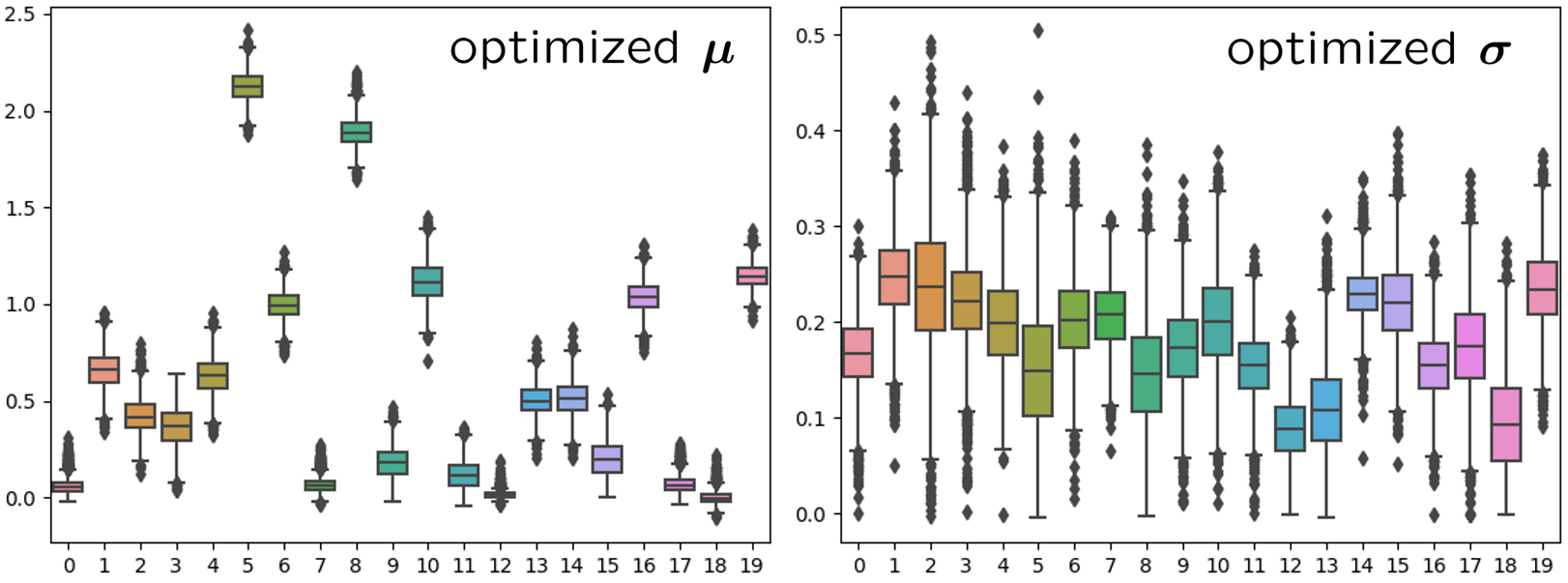}
     \caption{\textbf{Per-channel optimized statistics.} Distributions of the first $20$ channels of the optimized statistics of $\bs{\mu}$ (\textit{Left}) and $\bs{\sigma}$ (\textit{Right}). Each boxplot shows the interquartile range (IQR) that contains 50\% of the data: Its bottom and top edges delimit the first and third quartiles respectively. The horizontal line inside the box denotes the data median. The whiskers extend from the edges of the box to the furthest point within $1.5$ times the IQR, in each direction. Outlier points beyond these limits are individually plotted (diamonds).}
    \label{fig:learned_parameters_boxplots}
\end{figure}

\smallskip\noindent\textbf{Diversity of optimized statistics.}
To verify that the statistics --- optimized for the same number of iterations with the same $\prompt$ but from different starting anchor points $\emb{\src{\mb{f}}}$ --- are diverse, we show in~\cref{fig:learned_parameters_boxplots} the boxplots of optimized parameters on the first $20$ channels of $\stot{\mb{f}}$ (for prompt ``driving at night'').

\smallskip\noindent\textbf{Training from scratch on augmented features.} In \method, we start with a source-only trained model (\cref{algo:PODA_overall}, line 2) then we fine-tune it on augmented features (\cref{algo:PODA_overall}, line 5). This is the general setting for domain adaptation. However, since our method performs domain adaptation under the assumption of label preservation, we also experimented training the model from scratch on augmented features. The results (\cref{tab:from_scratch}) show the importance of the first, source-only training step.

\begin{table}[t]
	\setlength{\tabcolsep}{0.015\linewidth}
	\centering
	\begin{tabular}{lcccc}
		\toprule
		Method & Night & Snow & Rain & GTA5\\
		\midrule
        \method no src pretrain & 22.46 & 36.73 & 39.70 & 39.57\\
        \midrule
        \method& \textbf{25.03} & \textbf{43.90} & \textbf{42.31} & \textbf{41.07}\\
	    
  \bottomrule
	\end{tabular}
	\smallskip\caption{\textbf{Importance of source-only pre-training}. Semantic segmentation performance (mIoU \%) of \method \vs its variant without source-only training, when adapting from Cityscapes to ACDC Nigt/Snow/Rain and to GTA5.}
\label{tab:from_scratch}
\end{table}

\smallskip\noindent\textbf{Testing \method on other datasets.~}
\method does not use target datasets at any point in training. Although there is no reason for the improvements observed to be specific for the datasets we test on, we show in \cref{tab:more_night} the performance of the model adapted using ``driving at night'' on two additional night-time driving scenes datasets:
\begin{itemize}
    \item \textbf{Nighttime Driving~\cite{dai2018dark}} test set, which consists of 50 annotated images of night driving scenes, with resolution of $1920\times1080$.
    \item \textbf{NightCity~\cite{tan2021night}}, which is a large dataset of 4297 night-time driving scenes collected from many cities around the world; We tested on the validation/testing set, which consists of 1299 images of resolution $1024\times512$.
\end{itemize}

\begin{table}[t]
        \centering
\resizebox{1.0\linewidth}{!}{
	\setlength{\tabcolsep}{0.015\linewidth}
	\begin{tabular}{lccc}
		\toprule
		& ACDC Night &  \makecell{Nighttime\\Driving~\cite{dai2018dark}} & NightCity~\cite{tan2021night}\\
		\midrule
		  Source-only & 18.31\textcolor{white}{\vartn{0.00}} & 29.61\textcolor{white}{\vartn{0.00}} & 25.63\textcolor{white}{\vartn{0.00}}\\
        \midrule
	    \method & \textbf{25.03\vartn{0.48}} & \textbf{33.98\vartn{0.61}} & \textbf{28.90\vartn{0.61}}\\
  \bottomrule
	\end{tabular}}
	\smallskip\caption{\textbf{\method{} at night.} Segmentation performance (mIoU \%) of \method adapted from Cityscapes to nighttime with $\prompt$ = ``driving at night'', on three different night-time driving datasets.}
\label{tab:more_night}
\end{table}

\section{Class-wise performance}

\noindent We report class-wise IoUs in~\cref{tab:cityscapes_to_others}.

\begin{table*}[t!]
	\setlength{\tabcolsep}{0.007\linewidth}
        \renewcommand{\arraystretch}{1.2}
        \newcommand{\varclass}[1]{}
        \newcommand{\varmio}[1]{\vartn{#1}}
        \newcommand{\best}{\bf}
		\centering
		\resizebox{\linewidth}{!}{
	  \begin{tabular}{l c l c c c c c c c c c c c c c c c c c c c | c}
			\toprule
			\multicolumn{20}{c}{} \\
			Source
			& \makecell{Target\\eval.}
			& Method
			& \rotatebox{90}{road}
			& \rotatebox{90}{sidewalk}
			& \rotatebox{90}{building} 
			& \rotatebox{90}{wall} 
			& \rotatebox{90}{fence} 
			& \rotatebox{90}{pole} 
			& \rotatebox{90}{traffic light} 
			& \rotatebox{90}{traffic sign} 
			& \rotatebox{90}{vegetation} 
			& \rotatebox{90}{terrain} 
			& \rotatebox{90}{sky} 
			& \rotatebox{90}{person} 
			& \rotatebox{90}{rider} 
			& \rotatebox{90}{car} 
			& \rotatebox{90}{truck} 
			& \rotatebox{90}{bus} 
			& \rotatebox{90}{train} 
			& \rotatebox{90}{motorcycle} 
			& \rotatebox{90}{bicycle} 
			& mIoU\%\\
			\midrule

        \multirow{16}{*}{CS}& \multicolumn{22}{c}{\cellcolor{gray!30}$\prompt$ = ``driving at night''}\\
			& \multirow{3}{*}{\makecell{ACDC\\Night}} & source-only & 70.42 & 18.32 & \best43.83 & 6.11 & 17.08 & 23.52 & \best24.51 & 19.76 & 39.74 & 6.11 & 0.78 & 21.62 & 8.96 & 23.08 & 2.53 & 0.00 & 3.27 & 8.42 & 9.87 & 18.31\textcolor{white}{\varmio{0.00}}\\
    & & CLIPstyler & 73.96\varclass{} & 23.26\varclass{} & 42.16\varclass{} & 3.31\varclass{} & 7.21\varclass{} & \best35.49\varclass{} & 
    23.34\varclass{} & 19.01\varclass{} & \best45.41\varclass{} & 8.81\varclass{} & \best27.87\varclass{} & 21.06\varclass{} & 8.48\varclass{} & 38.17\varclass{} & 1.84\varclass{}
        & 0.00\varclass{} & 11.54\varclass{} & \best10.38\varclass{} & 4.89\varclass{} & 21.38\varmio{0.36} \\
    & & \method & \best77.54\varclass{}
    & \best26.90\varclass{1.09}
    & 42.71\varclass{}
    & \best13.51\varclass{}
    & \best21.36\varclass{}
    & 33.52\varclass{}
    & 23.70\varclass{}
    & \best21.73\varclass{}
    & 39.91\varclass{}
    & \best9.51\varclass{}
    & 19.40\varclass{}
    & \best28.80\varclass{}
    & \best11.85\varclass{}
    & \best50.89\varclass{}
    & \best10.14\varclass{}
    & 0.00\varclass{}
    & \best20.76\varclass{}
    & 8.76\varclass{}
    & \best14.50\varclass{}
    & \best25.03\varmio{0.48}\\
  
    \arrayrulecolor{gray!80}
    \cmidrule{2-23}

        & \multicolumn{22}{c}{\cellcolor{gray!30}$\prompt$ = ``driving in snow''}\\
        & \multirow{3}{*}{\makecell{ACDC\\snow}} & source-only & 70.47 & 23.50 & 63.80 & 17.96 & \best27.36 & \best38.52 & \best56.26 & \best45.00 & \best83.00 & 10.75  & 83.65 & 47.73 & 0.72 & 61.42 & 21.87 & 5.90 & 21.58 & 35.83 & 31.01 & 39.28\textcolor{white}{\varmio{0.00}}\\
    & & CLIPstyler& 74.29\varclass{}
& 31.25\varclass{}
& 69.17\varclass{}
& 15.21\varclass{}
& 25.21\varclass{}
& 36.83\varclass{}
& 44.79\varclass{}
& 42.56\varclass{}
& 76.87\varclass{}
& 11.07\varclass{}
& 91.48\varclass{}
& \best53.23\varclass{}
& 0.13\varclass{}
& 67.66\varclass{}
& 23.88\varclass{}
& \best9.14\varclass{}
& 36.48\varclass{}
& \best42.67\varclass{}
& 28.76\varclass{}
& 41.09\varmio{0.17}
\\
        & & \method & \best75.40\varclass{}
        & \best34.61\varclass{}
        & \best75.22\varclass{}
        & \best26.77\varclass{}
        & 27.34\varclass{}
        & 35.20\varclass{}
        & 52.68\varclass{}
        & 44.37\varclass{}
        & 82.01\varclass{}
        & \best14.16\varclass{}
        & \best93.72\varclass{}
        & 50.51\varclass{}
        & \best0.99\varclass{}
        & \best69.11\varclass{}
        & \best26.64\varclass{}
        & 2.72\varclass{}
        & \best46.98\varclass{}
        & 42.64\varclass{}
        & \best33.09\varclass{}
        & \best43.90\varmio{0.53}\\
    \cmidrule{2-23}
        &\multicolumn{22}{c}{\cellcolor{gray!30}$\prompt$ = ``driving under rain''}\\
		& \multirow{3}{*}{\makecell{ACDC\\rain}} & source-only & 74.10 & 31.98 & 63.07 & \best15.08 & \best23.92 & \best41.31 & \best50.12 & \best44.43 & 79.93 & 22.07 & 87.45 & 47.99 & 4.39 & 68.92 & 10.35 & 18.52 & 13.64 & 7.03 & 21.58 & 38.20\textcolor{white}{\varmio{0.00}} \\
  & & CLIPstyler & 73.71\varclass{}
  & 36.09\varclass{}
  & 68.91\varclass{}
  & 3.77\varclass{}
  & 16.99\varclass{}
  & 36.94\varclass{}
  & 39.75\varclass{}
  & 36.44\varclass{}
  & 78.21\varclass{}
  & 20.64\varclass{}
  & 91.79\varclass{}
  & 40.34\varclass{}
  & \best9.65\varclass{}
  & \best74.54\varclass{}
  & 13.16\varclass{}
  & \best20.33\varclass{}
  & 12.73\varclass{}
  & 14.06\varclass{}
  & 18.26\varclass{}
  & 37.17\varmio{0.10}\\
  
  & & \method & \best76.60\varclass{}
  & \best38.52\varclass{}
  & \best78.01\varclass{}
  & 15.02\varclass{}
  & 22.53\varclass{}
  & 40.33\varclass{}
  & 45.39\varclass{}
  & 41.40\varclass{}
  & \best86.85\varclass{}
  & \best37.97\varclass{}
  & \best96.46\varclass{}
  & \best50.39\varclass{}
  & 6.35\varclass{}
  & 74.19\varclass{}
  & \best19.19\varclass{}
  & 7.98\varclass{}
  & \best22.06\varclass{}
  & \best21.04\varclass{}
  & \best23.65\varclass{}
  & \best42.31\varmio{0.55}\\

    \cmidrule{2-23}
    &\multicolumn{22}{c}{\cellcolor{gray!30}$\prompt$ = ``driving in a game''}\\
    & \multirow{3}{*}{GTA5} & source-only & 68.72 & 22.65 & 78.79 & 36.81 & \best17.31 & 39.66 & \best39.33 & 14.84 & \best72.61 & 22.53 & 87.31 & 57.50 & 26.14 & 74.29 & \best44.57 & \best20.45 & 0.00 & 18.30 & 10.35 & 39.59\textcolor{white}{\varmio{0.00}}  \\
    & & CLIPstyler& 73.06\varclass{} 
    & \best29.89\varclass{}
    & 77.86\varclass{}
    & 25.50\varclass{}
    & 11.69\varclass{}
    & \best39.72\varclass{}
    & 35.88\varclass{}
    & \best24.04\varclass{}
    & 67.38\varclass{}
    & 12.75\varclass{}
    & 88.77\varclass{}
    & 46.58\varclass{}
    & 33.38\varclass{}
    & 72.03\varclass{}
    & 42.79\varclass{}
    & 11.12\varclass{}
    & 0.00\varclass{}
    & 28.84\varclass{}
    & \best14.61\varclass{}
    & 38.73\varmio{0.16}\\
    
    & & \method & \best73.93\varclass{}
    & 22.69\varclass{}
    & \best78.82\varclass{}
    & \best37.52\varclass{}
    & 14.17\varclass{}
    & 36.97\varclass{}
    & 33.14\varclass{}
    & 17.34\varclass{}
    & 72.44\varclass{}
    & \best26.22\varclass{}
    & \best88.85\varclass{}
    & \best62.69\varclass{}
    & \best37.04\varclass{}
    & \best74.33\varclass{}
    & 43.03\varclass{}
    & 11.91\varclass{}
    & 0.00\varclass{}
    & \best35.33\varclass{}
    & 13.91\varclass{}
    & \best41.07\varmio{0.48}\\
	\arrayrulecolor{black}
	\midrule
	&\multicolumn{22}{c}{\cellcolor{gray!30}$\prompt$ = ``driving''}\\
    \multirow{3}{*}{GTA5}& \multirow{3}{*}{CS} & source-only & 58.97 & 20.92 & 72.84 & 16.53 & \best24.58 & 31.37 & 34.77 & 23.62 & 82.12 & 17.04 & 66.28 & \best63.46 & 14.72 & \best81.27 & \best20.83 & 17.19 & 4.68 & \best20.57 & 19.56 & 36.38\textcolor{white}{\varmio{0.00}}  \\
    & & CLIPstyler& 66.70\varclass{} 
    & 23.63\varclass{}
    & 64.12\varclass{}
    & 5.08\varclass{}
    & 3.66\varclass{}
    & 20.67\varclass{}
    & 19.31\varclass{}
    & 18.10\varclass{}
    & 81.68\varclass{}
    & 12.36\varclass{}
    & \best81.04\varclass{}
    & 54.64\varclass{}
    & 0.52\varclass{}
    & 73.47\varclass{}
    & 20.65\varclass{}
    & \best22.30\varclass{}
    & 4.03\varclass{}
    & 15.79\varclass{}
    & 10.73\varclass{}
    & 31.50\varmio{0.21}\\
    
    & & \method & \best84.34\varclass{}
    & \best36.73\varclass{}
    & \best79.43\varclass{}
    & \best18.33\varclass{}
    & 16.54\varclass{}
    & \best36.93\varclass{}
    & \best38.45\varclass{}
    & \best33.81\varclass{}
    & \best82.44\varclass{}
    & \best19.14\varclass{}
    & 75.90\varclass{}
    & 62.65\varclass{}
    & \best16.47\varclass{}
    & 75.48\varclass{}
    & 15.68\varclass{}
    & 19.57\varclass{}
    & \best11.28\varclass{}
    & 16.53\varclass{}
    & \best21.76\varclass{}
    & \best40.08\varmio{0.52}\\
    \bottomrule
		\end{tabular}}
	\smallskip\caption{\textbf{Zero-shot domain adaptation in semantic segmentation.} Performance (mIoU\%) of~\method~compared against~CLIPstyler~\cite{kwon2022clipstyler} and source-only baseline. Results are grouped by source domain and target domain with associated {\setlength{\fboxsep}{1pt}\colorbox{gray!30}{$\prompt$}}. CS stands for Cityscapes~\cite{cordts2016cityscapes}. This table provides details of the main results in \cref{tab:main_results}.}
	\label{tab:cityscapes_to_others}
\end{table*}

\section{\method{} for Object Detection}
Here, we share the implementation details for our object detection experiments (\cref{sec:extensiontasks} and \cref{tab:poda_od_results}). 
We used the implementation of Faster R-CNN~\cite{ren2015faster} from the MMDetection library.\footnote{\url{https://github.com/open-mmlab/mmdetection}}
With Cityscapes as source dataset, we trained all models for $8$ epochs using the SGD optimizer with $0.9$ momentum and $1\text{e}{-4}$ weight decay.
The initial learning rate $lr$ is set as $1\text{e}{-2}$ and is dropped by a factor of 10 after the $7$th epoch; the same $lr$ scheme is used in source only and~\method{} trainings.
With Day-Sunny split of the DWD dataset as source, models are trained for $20$ epochs using a similar SGD optimizer.
When training on source, the learning rate starts at $1\text{e}{-3}$ and drops at the $9$th epoch to $1\text{e}{-4}$; in~\method{} training, the learning rate is ten times less.

{\small
\bibliographystyle{ieee_fullname}
\bibliography{egbib}
}

%%%%%%%%% REFERENCES

\end{document}